%% file: main.tex
\documentclass{article}
\usepackage{spconf,amsmath,graphicx}

\usepackage{hyperref}
\usepackage{color}
\definecolor{cadmiumgreen}{rgb}{0.0, 0.42, 0.24}

\newcommand{\ie}[0]{\textit{i.e.},~}

\usepackage{nomencl}
\usepackage{subfigure}
\makenomenclature
\usepackage{multirow}
\usepackage[table,xcdraw]{xcolor}
\usepackage{color,soul}
\DeclareMathOperator*{\poll}{\textit{MAX}}
\usepackage{multirow}

\usepackage{amssymb}
\usepackage{hyperref}
\usepackage{comment}
\hypersetup{
    colorlinks=black,
    linkcolor=black,
    filecolor=black,      
    urlcolor=magenta,
    citecolor= black,
}
\usepackage{float}
\usepackage{comment}
\usepackage{atbegshi,picture}
\usepackage{fancyhdr}
\usepackage{eso-pic}% http://ctan.org/pkg/eso-pic

\title{SPATIO-TEMPORAL GRAPH-RNN FOR POINT CLOUD PREDICTION}
\name{Pedro Gomes,  Silvia Rossi, and Laura Toni}
\address{ Department of Electronic \& Electrical Engineering \\ University College of London, UK\\
	email:\{pedro.gomes.19, s.rossi, l.toni\}@ucl.ac.uk}
\AtBeginShipoutNext{\AtBeginShipoutUpperLeft{%
  \put(\dimexpr\paperwidth -20.5cm\relax,-1.5cm){\begin{minipage}{555pt}{© 2021 IEEE. Personal use of this material is permitted. Permission from IEEE must be obtained for all other uses, in any current or future media, including reprinting/republishing this material for advertising or promotional purposes, creating new collective works, for resale or redistribution to servers or lists, or reuse of any copyrighted component of this work in other works.} \end{minipage} } } 
}

\begin{document}

\maketitle
\begin{abstract}
In this paper,  we propose an end-to-end learning network to predict future frames in a point cloud sequence. As main novelty, an initial layer learns topological information of point clouds as geometric features, to form representative spatio-temporal neighborhoods.
This module is followed by multiple Graph-RNN cells. Each cell learns points dynamics (i.e., RNN states) by processing each point jointly with the spatio-temporal neighbouring points.  
We tested the network performance  with a MINST dataset of moving digits, a synthetic human bodies motions and JPEG dynamic bodies datasets. Simulation results demonstrate that our method outperforms baseline ones that neglect geometry features information. %is able to accurately group points and make correct predictions showing its ability to model short- and long-term relationships while preserving the spatial structure. 
%The code has been made available at \href{https://github.com/pedro-dm-gomes/Graph-RNN}{https://github.com/pedro-dm-gomes/Graph-RNN}.
\end{abstract}

\section{Introduction}
Point clouds (PCs) sequences provide a flexible and rich geometric representation  of  volumetric  content,  quickly  becoming  an  attractive representation for applications such as autonomous driving~\cite{nuscenes},  mixed reality application services~\cite{VR}, cultural heritage \cite{Culture_3D}.  This has motivated intense research  toward PC processing, with strong focus  on \emph{static} PCs, leaving the \emph{dynamic} PC processing (DPC) usually overlooked. In this work, we focus on DPC processing and specifically  on the prediction of point cloud dynamics. Namely, given PC frames $P_1, P_2, \ldots, P_t$ we are interested in predicting $\hat{P}_{t+1}$, \emph{with no prior knowledge} on the ground truth  $P_{t+1}$.

In the current literature, DPCs processing has been approached from two overlapping directions: (1) motion estimation (ME) and motion compensation (MC) for PC compression; (2) 3D motion flow prediction (usually deep-learning based) for high-level tasks (e.g., gesture recognition). %\footnote{add the part commented?? \notePG{don't think is needed}%\blue{This is highly important in intelligent systems, in which decisions need to be taken based on future prediction, but also in coding schemes as they can reconstruct PC without motion vector information being sent.} }. %The former is adopted to compress PCs  by leveraging on the temporal redundancy between sequential PC frames.  The latter is  slightly different approach,    in which deep learning processing is considered for 3D motion flow prediction.
 Both approaches share  a common challenge: extraction of temporal correlations between sequential PC frames, challenged by the irregular structure and by the lack of  explicit point-to-point correspondence.   At the same time, these two directions have fundamentally different goals and setups: the former aimed at extracting the
%It is worth saying that the motion flow prediction is   more challenging than ME and MC. The reason being that we do not only need to extract the
motion vector from two \emph{known} consecutive frames,  the latter focused on a much more challenging task of   prediction of future \emph{unknown} PC frames. This requires  learning both the short- and long-term PC trajectory. 
Another key difference lies in the developed solutions: ME mainly addresses the lack of correspondence  either by  projecting  the 3D PC into the 2D domain and adopting  mature algorithms from 2D video compression~\cite{V-PCC} or by developing  3D ME methodologies, preserving the volumetric information of the PCs~\cite{mekuria2016design,de2017motion,DBLP:journals/corr/ThanouCF15}.

Motion flow prediction  involves deep learning processing instead, the irregular and unordered structure of PC prevents the immediate adoption of convolution neural networks. Within this framework,  PointNet \cite{qi2017pointnet}  has become a pillar work for static PC processing, capable of learning directly from raw PC data with no pre-processing: each point in the PC is processed independently and all point features are aggregated subsequently. Modeling points independently achieves permutation invariance, but at the price of losing the geometric relationship between points, a key piece of information in PCs that we aim at retaining.  %and extensions have been proposed for dynamic PCs~\cite{CloudLSTM,PointLSTM,PointRNN}), 
%The PointNet architecture was extended to capture motion as features from spatio-temporal neighborhoods in ~\cite{FlowNET3D, FlickerNet}. These methods, however, merely focus on short-term modeling and have insufficient abilities to capture long-term relationships.
To learn the dynamic behavior of sequential data, recent works% have adapted recurrent neural networks (RNNs)  to PC processing
~\cite{CloudLSTM,PointLSTM,PointRNN}
has extended PointNET architecture to recurrent neural networks (RNNs), predicting the 3D motion flow of PCs. In the PointRNN model~\cite{PointRNN}, for example,  each point is processed individually by RNN cells with the output being the point state (i.e., the motion of the point). Each point state is extracted by aggregating state information from neighboring points.  The neighborhood of a point of interest is defined as the $k$ nearest neighbor ($k$-nn) points in the previous frame, where the proximity is measured based on the  point coordinates. 
This methodology inherits the ability to capture the dynamic behavior of sequential data from RNN models, as well as permutation invariance from PointNet architecture. However, it suffers from the same shortcoming of PointNet: \emph{lack of geometric relationship between points} which may  lead  to \emph{i)} loss of structure during PC reconstruction; \emph{ii)} poor $k$-nn neighborhood as grouping points only based on coordinates might connect two points close in space but not belonging to the same segment, hence not sharing the same motion. %Interestingly, the powerful idea of exploiting    the geometry information of the PC during the processing is   well accepted in the community  for static PC processing~\cite{DBLP:DGCNN, Graph-denoising, Graph-CNN}, or for ME~\cite{DBLP:journals/corr/ThanouCF15}, but not for motion flow prediction.  

%Furthermore   grouping points based only on coordinates is not ideal. In some scenarios, this can lead to the grouping of points that while still close in  space, might belong to completely different segments without temporal correlations between them.
%One of the consequences for example the loss of structure in the PC reconstruction in~\cite{PointRNN}. Recently, graph representations of PCs have gained popularity, due to their inherent strong capability to handle irregular data. Graph-based processing has achieved significant performance in PCs analysis, \textit{e.g}., representation, denoising, classification tasks \cite{DBLP:DGCNN, Graph-denoising, Graph-CNN}. These works however, have been mainly focus on \emph{static} PCs, leaving the dynamic counterpart usually overlooked.

In this paper, we seek to bridge the gap between graph-based representations of PC~\cite{DBLP:DGCNN, Graph-denoising, Graph-CNN} and deep learning motion flow prediction. % Inspired by the graph-based processing for static PCs~\cite{DBLP:DGCNN, Graph-denoising, Graph-CNN}, 
%We improve PointRNN \cite{PointRNN} by designing a more effective spatio-temporal correlation by incorporating geometric information between points. Inspired by the graph-based processing for static PCs~\cite{DBLP:DGCNN}, we aim at learning topological information of PC as geometric features, and leverage on those learned features to form more representative local neighborhoods of points. This results in spatio-temporal aggregation of points that share a common local geometry instead of just points close in space.  
We propose an end-to-end architecture, where an initial pre-processing step learns topological information of PC as geometric features, and leverage on those learned features to form more representative local neighborhoods of points than PointRNN models. From the learned features, a Graph-RNN constructs spatio-temporal $k$-nn graphs. This results in spatio-temporal aggregation of points that share  common local features instead of only points coordinates.   The temporal correlations from the spatio-temporal graph are aggregated to learn point states. The Graph-RNN learns points states, which retain the model dynamic information over time and allow to model long-term point trajectory.   The proposed solution has been validated on moving MNIST point cloud dataset used in the literature~\cite{PointRNN} as well as on a synthetic human b odies motions and JPEG dynamic bodies datasets~\cite{Irene, JPEG_Pleno}. Simulation results demonstrate that our method can make correct PC predictions showing its ability to accurately group points and model long-term relationships while preserving the spatial structure.

%\hlp{While these works extract instead of predicting the motion flow, failing in supporting long-term motion prediction, some of these works are based on the idea of exploiting the geometry information of the PC during the processing}~\cite{DBLP:journals/corr/ThanouCF15}\hl{[REf]}. \hlp{Powerful idea  well accepted in the community mainly for static PC processing}~\cite{DBLP:DGCNN, Graph-denoising, Graph-CNN},\hlp{which we aim to extend to the the dynamic PC processing counterpart.}

%In summary the main contributions of this paper are:
%\begin{itemize}
%
 %   \item Design  a spatio-temporal correlation able to effectively extract temporal correlations from irregular point cloud sequence data by leveraging  learned  features,  describing the local topology, to aggregate similar points across frames. 
  %  
   % \item A Graph-RNN that can leverage the spatio-temporal correlation to model long term relationship while preserving spatial structure.
    %
    %\item Evaluation of the propose model in prediction task on human bodies motions, demonstrating the ability to  make accurate predictions.
%\end{itemize}

%%%%%%%%%%%%%%
\section{Proposed Method}
\label{sec:method}
\begin{figure*}[t]
 \centering
  \includegraphics[width=0.85\textwidth,scale =0.6]{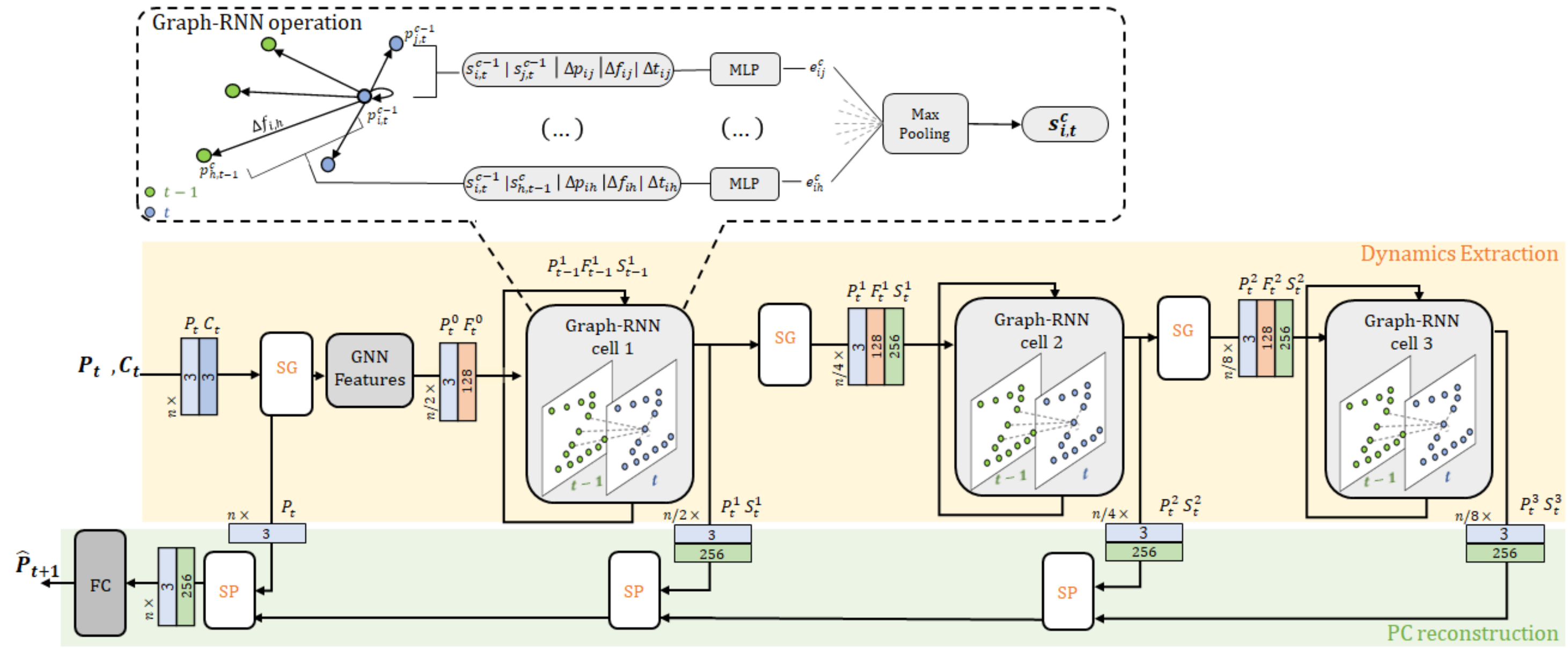}
  \caption{Scheme of the complete hierarchical architecture, composed of four main components: Sampling and Grouping (SG); GNN for Features Learning ; Graph-RNN (diagram of it's operation included in dashed area) ; States propagation (SP). }
  \label{scheme}
\end{figure*}
We denote a point cloud frame consisting of $n$ points by $P_t =\{p_{1,t}, p_{2,t}, \ldots p_{n,t}  \} $ with  $p_{i,t}\in \mathbb{R}^3$ being the euclidean coordinates of    point $i$ in $P_t$. Each PC has additional attributes (\ie point color) denoted by  $C_t =\{c_{1,t}, c_{2,t}, \ldots c_{n,t} $, with  $c_{i,t}\in  \mathbb{R}^3$ the associated color component.
Given a point cloud sequence $\mathcal{P}=(P_{1}, P_{2}, ..., P_{T})$ composed by $T$ frames and additional attributes $\mathcal{C}=(C_{1}, C_{2}, ..., C_{T})$, our  goal is to predict the coordinates of future point clouds  $\hat{P}_{T+1},\ldots,\Hat{P}_{T+Q}$, with $Q$ being the prediction horizon. 

To reach this goal, we proposed an interactive framework (Fig.~\ref{scheme}), which allows us to predict  future trajectories of points via RNN cells.  At each iteration,  the network processes one input frame $P_{t}$ and its color attribute $C_{t}$ giving as output the prediction of the successor frame $\hat{P}_{t+1}$. The architecture is composed of two phases: \emph{i)}  a dynamics extraction (DE) phase where the PC dynamic behaviour is captured in the form of point states, \emph{ii)} a PC reconstruction phase where the states are concatenated and used to output the PC prediction.
In the DE phase,  as key novelty, we  pre-process  the point cloud  to extract point features that carries  local geometry information. Specifically, a initial graph neural network (GNN) module transforms the 3D space into an higher dimensional feature space and sends the output to a Graph-RNN cell. In each cell, each point is processed independently to preserve permutation invariance. Specifically, each point state is extracted by aggregating information from its $k$-nn neighborhood. After the  Graph-RNN cells, the PC reconstruction phase begins. The states are propagated and processed by a fully connected layer (FC) to estimate motion vectors, used to predict the next frame $\hat{P}_{t+1}$.
Before each Graph-RNN cell, the point cloud is down-sampled. It is then  up-sampled to its original size before the final FC layer. The down-sampling and up-sampling blocks are implemented as in \cite{qi2017pointnet++} and we refer the readers to Section A of the Appendix.
%~\cite{Graph-RNN} 
for further information. The intuition for the design hierarchical architecture is to learn  states at  multiples scales:  the first Graph-RNN cell handles a dense PC and learns states in local regions (corresponding to local motions), while  the last Graph-RNN cell   learns states in a sparser PC with more distant points included in the   neighborhood (corresponding to more global motions).%
%To this end, Not being a key novelty of this paper, and for  the sake of brevity, in the following we provide a short description of the blocks   and we refer the reader to \cite{qi2017pointnet++} for a detail description. The sampling and grouping  (SG)  module takes a PC  with $n$ points and uses the farthest point sampling (FPS) algorithm to sample $l$ points. %\emph{The sampled points are defined as centroids of local regions. Each region is composed of the $k$ closest neighborhood points to the centroid point. The features and states of the points in a region are max pooled into a single feature and state representation. This representation becomes the feature and the state of the centroid point. The SG module outputs the $n'$ sampled points and their updated feature and state.} \lt{I would skip this part. you have yet another graph and this part is anyway not a contribution - I would leave a more technical description in the appendix. Then we will upload on Arxiv the paper and here we reference the Arxiv. }
%Once the PC dynamics is extracted \lt{in terms of state?}\notePG{yes}, the point cloud needs to be upsampled to reconstruct the future PC frame. \lt{why you upsample and downsample? give intuition here on the general description above} A states propagation (SP) module will then upsample the states representations associated with the intermediate points $l$ to the original points $n$.
%

We now provide more details on the key modules that are part of our contributions:   GNN-based pre-processing and Graph-RNN cells. % The model also includes several auxiliary processing modules named \emph{Sampling and Grouping} (SG) and \emph{States Propagation} (SP). 

\subsection{GNN for Feature Learning}
\label{sec:GNN}
Given $P_t$ and $C_t$ as input, we construct a directed $k$-nn \emph{coordinate graph} $\mathcal{G}^{C}_t=(P_t,\mathcal{E}^C_t)$ with   vertices $P_t$ and edges  $\mathcal{E}^C_t$. Each edge connects a point to its $k$-nearest neighbors based on euclidean distance. The graph includes self-loop, meaning each point is also connected to itself.  Given the coordinate graph as input,  the GNN module  learns the geometric features $F_t \in \mathbb{R}^{ n \times d_f}$.  %From \red{[Bronstein]}, we learn the features by taking into account the relationship of each point from its neighboring \lt{give intuition: does it mean that points with similar features have the same neighborhood? }  Inspired by \red{[Broinstein]}, we learn the point features not directly from their embeddings, but rather from 
The GNN is composed of $L$ layers, and  in each layer features are learned by aggregating information along the edges. %\lt{is this the case of aggregation? this usually means that things are averaged along the edges... don't look to tbe the case for me.}

Inspired by  \cite{DBLP:DGCNN}, we learn the features by taking into account the relationship (similarity) between neighboring points. At the  $l$-th layer of the GNN, the \emph{edge features} $e^l_{i,j}$ are learned  for each point $i$ and for each neighboring node $j$.  
% the \emph{edge feature} $e^l_{i,j}$ are learned for each neighboring node $j$. 
This is  done by concatenating the  input point feature $f_{i,t}^{l-1}$ and the point coordinates $p_{i,t}$, with the geometry and color displacement/difference  between the points $i$ and  $j$ ($\Delta p_{ij}$, $\Delta c_{ij}$, respectively). We then apply a symmetric aggregation operation on the edge features associated with all the edges emanating from each point.  More formally, the edge features ($e^{l}_{i,j}$) and the output point features  ($f^{l}_i$) are  obtained as follows:
\begin{align}
     e^{l}_{i,j} &= h_F^l(f_{i,t}^{l-1} \; ; p_{i,t} \; ; \Delta p_{ij} \; ; \Delta c_{ij} )\\
     f^{l}_i &=  \poll_{ j:(i,j) \in \mathcal{E}^C} \big\{ e^{l}_{i,j} \big\}
\end{align}
where \(h_F\) is a nonlinear learnable function that can be implemented with a multi layer perceptron (MLP), '$;$' identifies the  concatenation operation and \(\poll\) represents the element-wise max pooling function.
Note that for the first layer $l=1$, we set  $f_{i,t}^{0}$ as a null entry and the output of the $L$-th layer is the geometric feature  $F_t = [f_{1,t}^{L}, f_{2,t}^{L}, \ldots, f_{n,t}^{L}]$. 

%Overall the GNN operation is spatial convolution, that convolve a central point representation with its neighbors’ representations to produce a updated representation of the central point capturing the local topology.

%\lt{not clear - have you introduced f before?? Also in classical GNN you have a weighted sum, is this the case?}\notePG{no, there is no attention or weights, the only weighted aspect is inserted by the concatenation, it that aspect is the same as broinstein } \lt{ Are you using the broinstein GNN or yours?}\notePG{mine, the difference being i dont updated the graph between layers, and the values chosen for concatenation }\lt{ it could be good to add more details here. for example in classical GNN you would have something like equation (5) here https://arxiv.org/pdf/2003.11702.pdf while you also state the dependency but not the actual operations that happen at the node level}.\notePG{Using that paper notation is good when there is no concatenation and we have attention network, only the features are processed, the concatenation is really what we show in the equation}\\

%%%%%%%%%%%%%%
\subsection{Graph-RNN }
\label{sec:GraphRNN}
Each Graph-RNN cell $c$ receives  the feature   $F_t^{c-1}$ and $P_t^{c-1}= [p^{c-1}_{1,t}, p^{c-1}_{2,t}, \ldots, p^{c-1}_{n,t}]$ as  input, with  $F_t^0$ being the output of the previous GNN module.  Given it iterative nature, the Graph-RNN cell takes into account the input and also its own output   $(P^{c}_{t-1},F^{c}_{t-1},S^{c}_{t-1})$ calculated at  the previous interaction $(t-1)$.  The cell  extracts the inner state $S_t^c = [s^c_{1,t}, s^c_{2,t}, \ldots, s^c_{n,t}] \in \mathbb{R}^{n\times d_s}$, with  $s^c_{i,t}$ being the state of  point $p^c_{i,t}$, representative of the point dynamic behavior. The new state is added to the unchanged coordinates $P^{c-1}_t$ and features $F^{c-1}_t$ and outputted as $(P^{c}_{t},F^{c}_{t},S^{c}_{t})$.   
Similarly to~\cite{PointRNN}, we consider three sequential Graph-RNN cells.

The  Graph-RNN operation is the depicted in Fig.~\ref{scheme} (dashed box). As first step, we compute a \emph{spatio-temporal feature graph} $\mathcal{G}^F_t$, in which each point is connected to $k$ nearest neighbors based on the feature distance. Specifically, for each input point $p_{i,t}^{c-1}$,  we compute the pairwise distance between $f_{i,t}^{c-1}$  and  features of other points $f_{j,t}^{c-1}$ (features input) and  $f_{j,t-1}^{c}$ (features from points in the past PC).  We force our implementation to take the equal number of points $k$ from $P^{c-1}_t$ as from  $P^{c}_{t-1}$ to avoid a one-side selection.
%of the other points both in the input  $f$ For each point, we evaluate the pairwise distance matrix in feature space from both  $F^{c-1}_t$ and $F^{c}_{t-1}$.  $f^{c-1}_{i,t}$ in $F^{c-1}_t$ 
In details, this is a spatio-temporal graph since each point is connected to points in the same PC (\emph{spatial relationship}) and points in the past PC (\emph{temporal relationship}).  
%
%First for every feature $f^{c-1}_{i,t}$ in $F^{c-1}_t$ we As compute a pairwise distance matrix in feature space from both  $F^{c-1}_t$ and $F^{c}_{t-1}$. A spatio-temporal graph is then constructed by taking the $k$ closest points based on feature distance. We force our implementation to take the equal number of points $k$ from $P^{c}_t$ as from  $P^{c}_{t-1}$ to avoid a one-side selection.
%
%
%We name the spatio-temporal graph \emph{feature graph} $\mathcal{G}^F_t$.  
Once the features graph is constructed, we learn edge features similarly to the GNN module. For the edge $(i,j)$, we concatenate the state of point $i$ ($s_i$), the state of point $j$ ($s_j$), and the coordinate, the feature and the time displacement ($\Delta_{p_{i,j}},\Delta_{f_{i,j}}\Delta_{t_{i,j}}$) between the two points. The concatenation is then processed by a shared MLP $(h_S)$. All edge features are then max pooled to a single representation into the update state $s_{i,t}$. Formally, 
\begin{align}
     e^{c}_{i,j} &= h_S^c (s^{c-1}_{i,t} ; s^{c'}_{j,t'} ; \Delta p_{ij} ; \Delta f_{ij} ;\Delta t_{ij} )\\
     s^{c}_{i,t} &=  \poll_{ j:(i,j) \in \mathcal{E}^F} \big\{ e^{c}_{i,j} \big\} 
\end{align}
%The graph is constructed based on the feature euclidean distance. Since the features characterize the points in the cloud, similar points will have a similar feature and therefore their relative feature distance is small. In this way, the use of feature distances allows to group points that represent the same part of the object across frames. 
When learning output states $S^{c}_t$, the Graph-RNN  cell considers the states in the previous frame $S^{c}_{t-1}$. This means that the network learns point movements taking into consideration the previous movements of points, allowing the cell to retain temporal information. The states act as a memory retaining  the history of movements and enabling for  network to model long-term relationships over time. 

%\notePG{Hey, Laura, ty for working on this so late on a friday,\\ Also the deadline is 14 (Sunday) so I as thinking of put in the results at 150k interactions and submit tonight to the arvix so we get the reference to use}

%%%%%%%%%%%%%%%%%%%%%

\subsection{Training}
\label{sec:training}
The architecture in Fig.~\ref{scheme} has multiple learnable parameters (in GNN, Graph-RNN, FC), which are end-to-end trained. We consider a supervised learning settings in which the loss function  relates to the prediction error between ground truth point cloud $P_t$ and  the  predicted one $\hat{P}_t$.  To evaluate the prediction error, we adopt the Chamfer Distance (CD) and Earth Moving Distance (EMD)  between $P_t$ and  $\hat{P}_t$ evaluated as follows~\cite{urbach2020dpdist}: 
\begin{align}
    d_{CD}(P, \hat{P} ) &= \sum_{p \in P} \min_{ \hat{p} \in \hat{P}} || p - \hat{p}||^2 + \sum_{p \in \hat{P}} \min_{ p \in P} || \hat{p} - p||^2 \\
    d_{EMD}(P, \hat{P} ) &= \min_{\theta:P \xrightarrow{} \hat{P} } \sum_{p \in P} || p - \theta(p) ||^2  
\end{align}
\noindent where \( \theta: P \xrightarrow{} \hat{P} \) is a bijection. The loss function used to train the network then is given by the sum of CD and EMD distances, namely $   \mathcal{L}(P, \hat{P}) =  d_{CD}(P, \hat{P} ) + d_{EMD}(P, \hat{P} )$.

%%%%%%%%%%%%%%% % 

\section{Experiments}
\label{sec:Experiments}
We implemented  the end-to-end network described in Sec.\ref{sec:method} in the case of $L=3$ layers within the GNN module and $C=3$ RNN cells\footnote{The code has been made available at \href{https://github.com/pedro-dm-gomes/Graph-RNN}{https://github.com/pedro-dm-gomes/Graph-RNN}.}. %\red{Given a sequence with $T$ frames (dependent of the dataset).} 
We consider both  short-term and  long-term prediction, with the former predicting only one future frame $Q=1$ (ground truth frame $P_t$ is used  to predict the next frame $\hat{P}_{t+1}$) while the latter predicting $Q= T/2$ future frames with  $\hat{P}_t$ being used  to predict the next frame $\hat{P}_{t+1}$.
As baseline models we consider: (1) \textit{Copy Last input} model which simply copies the past PC frame  instead of predicting it; (2) PointRNN ($k$-nn) model~\cite{PointRNN}, which neglects geometry information. 
In our experiments, we considered the following datasets:~\footnote{We provide detailed information in the supplementary information 
in (sec B) of the Appendix.} %\cite{Graph-RNN} }\\
\noindent\textbf{Moving MNIST Point Cloud}, created by converting the MNIST dataset of handwritten digits into moving point clouds, as in~\cite{PointRNN}, each sequence contains 20 ($T$) frames with either 128 (1 digit) or 256 points (2 digits) .\\
\noindent\textbf{Synthetic Human Bodies Activities}, synthetically generated by us  following~\cite{Irene} using the online service Mixamo \cite{Mixamo}\footnote{Due to copyright restriction imposed by Mixamo, we cannot provide the  dataset publicly available.  } in combination with the 3D animation software Blender \cite{Blender}. \\
\noindent\textbf{JPEG Pleno 8i Voxelized Full Bodies}, four benchmark sequences: longdress, loot, redandblack, and soldier \cite{JPEG_Pleno}. 
\\
In the two last datasets, each PC sequence contains 12 ($T$) frames and is downsampled to $4,000$ points. The network is trained with the Synthetic Human Bodies Activities dataset, which provides different levels of movements  (walking, jumping, dancing, etc) and tested on both datasets.

\begin{figure}[t!]
    \centering
    \centerline{\includegraphics[width=0.80\linewidth]{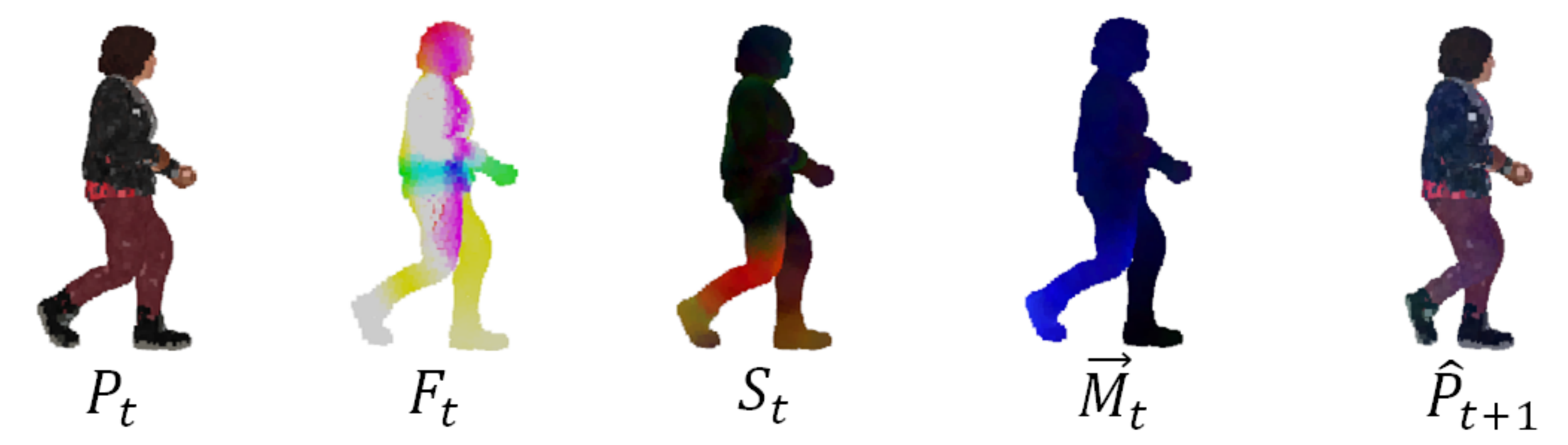}}
    \caption{ Multiple representation steps of short-term prediction Graph-RNN on Bodies dataset}
    \label{fig:r_bodies}
\end{figure}

\begin{figure}[t!]
    \centering
    \centerline{\includegraphics[width=1\linewidth]{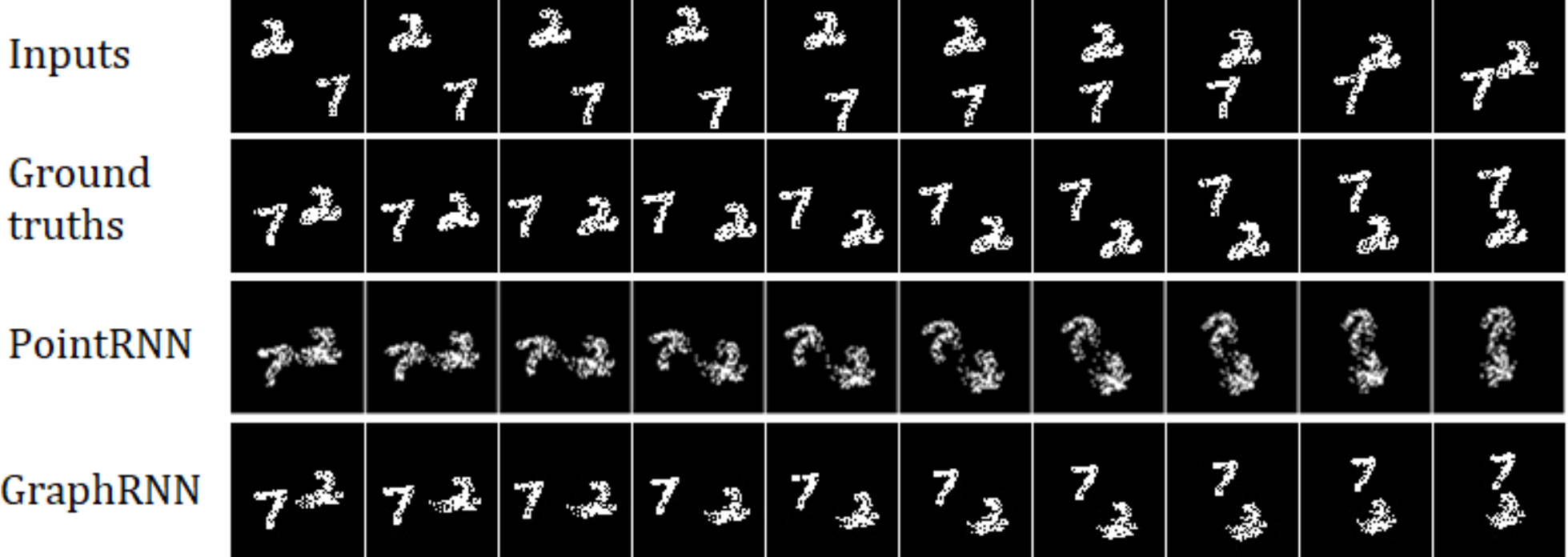}}
    \caption{Visualization of moving MNIST point cloud prediction with two moving digits (Hierarchical architecture)}
    \label{fig:results_mnist}
\end{figure}

To better understand our system, we visualized the learned features for one PC from the Synthetic  Human  Bodies dataset. Fig.~\ref{fig:r_bodies} depicts in sequence: the point cloud, the learned features $F_t$, the output state $S_t$, the reconstructed motion vector $M_t$, and the predicted PC. Principal Component Analysis (PCA) is used for the features visualization.  It is worth noting that  features can  segment the PC into regions sharing similar topology (leading to meaningful neighborhood in the features graph) and  states are able to capture the movement of moving parts --e.g.,  leg and foot taking a step forward. The states are directly translated into   motion vectors, used to make accurate prediction of the future frame.
A more complete comparison can be deduced from Fig.~\ref{fig:results_mnist},  depicting resultant and ground truth  PCs for the MINST dataset. Interestingly,  the predicted digits are   sharper and   clearer  in the Graph-RNN prediction than the PointRNN. This demonstrates that while both models capture the correct motion, the Graph-RNN is better at preserving the spatial structure over time. This is a direct effect from learning geometric features. %Can also be noted that the hierarchical architecture achieves significant better results compared with the basic architecture, and the use of color allows to learn more descriptive features resulting significant improvement

We now provide more quantitative results for both the MNIST dataset (Table 1), and  the Synthetic Human Bodies and JPEG  datasets (Table 2). For all datasets, the proposed Graph-RNN outperforms the PointRNN solution as well as the baseline solutions. From Table 1, it is worth noting that the hierarchical implementation (in which PC is sampled between RNN-cells) leads to a better prediction compared to the ``Basic" (not down-sampled) counterpart. This is expected as the hierarchical structure learns states at different resolution.  Finally,  the model ``Graph-RNN (color)" considers the color attributes when learning features, resulting in a more meaningful spatio-temporal neighborhood \cite{graph_color} and therefore in a better prediction.


\begin{table}[t!]
\centering
\resizebox{0.7\linewidth}{!}{
\begin{tabular}{|c|c|cc|cc|}
\hline
\multicolumn{6}{|c|}{MINST}                                                                                                                                                                    \\ \hline
\multicolumn{2}{|c|}{}                                            & \multicolumn{4}{c|}{Long-term Prediction}                                                                                  \\ \cline{3-6} 
\multicolumn{2}{|c|}{}                                            & \multicolumn{2}{c|}{One Digit}                              & \multicolumn{2}{c|}{Two digit}                               \\ \cline{3-6} 
\multicolumn{2}{|c|}{\multirow{-3}{*}{Method}}                    & CD                           & EMD                          & CD                            & EMD                          \\ \hline
\multicolumn{2}{|c|}{Copy last input}                             & 262.46                       & 15.94                        & 140.14                        & 15.8                         \\ \hline
                               & PointRNN                         & 5.86                         & 3.76                         & 22.12                         & 7.79                         \\
\multirow{-2}{*}{Basic}        & \cellcolor[HTML]{EFEFEF}GraphRNN &2.43 \cellcolor[HTML]{EFEFEF} & 2.40 \cellcolor[HTML]{EFEFEF} & 13.66 \cellcolor[HTML]{EFEFEF} & 6.13 \cellcolor[HTML]{EFEFEF} \\ \hline
                               & PointRNN                         & 2.25                         & 2.53                         & 14.54                         & 6.42                         \\
\multirow{-2}{*}{Hierarchical} & \cellcolor[HTML]{EFEFEF}Graph-RNN &  \textbf{1.22} \cellcolor[HTML]{EFEFEF} & \textbf{1.86} \cellcolor[HTML]{EFEFEF} &  \cellcolor[HTML]{EFEFEF} \textbf{4.62}  & \cellcolor[HTML]{EFEFEF}\textbf{3.97} \\ \hline
\end{tabular}}
\caption{ Prediction error of PointRNN and Graph-RNN with $k$-nn on the moving MNIST point cloud dataset.}
\end{table}

\begin{table}[t!]

\centering
\resizebox{1.0\linewidth}{!}{
\begin{tabular}{|c|cc|cc|cc|cc|}
\hline
                                                                                & \multicolumn{4}{c|}{ Synthetic Human Bodies}                                                      & \multicolumn{4}{c|}{JPEG Dynamic Bodies}                                         \\ \cline{2-9} 
                                                                                & \multicolumn{2}{c|}{Short-Term } & \multicolumn{2}{c|}{Long-Term } & \multicolumn{2}{c|}{Short-Term } & \multicolumn{2}{c|}{Long-Term } \\ \cline{2-9} 
\multirow{-2}{*}{\begin{tabular}[c]{@{}c@{}}Method\\ Hierarchical\end{tabular}} & \multicolumn{1}{c|}{CD}        & EMD       & \multicolumn{1}{c|}{CD}       & EMD       & \multicolumn{1}{c|}{CD}        & EMD       & \multicolumn{1}{c|}{CD}       & EMD       \\ \hline
Copy Last Input                                                                 & 0.161                              & 0.153         &    0.247                           &    0.408        &0.0004                                &0.029           & 0.0020                              &0.058           \\
\rowcolor[HTML]{EFEFEF} 
PointRNN                                                                        & \textcolor{black}{0.007}                                 & \textcolor{black}{0.104}           &0.066                               & 0.257           & \textcolor{black}{0.0005}                             &\textcolor{black}{0.034}           &0.0024                               & 0.082           \\
Graph-RNN                                                                        &\textcolor{black}{0.005}                                   & \textcolor{black}{0.078}             &0.077                               & 0.248          & \textcolor{black}{0.0003}                               & \textcolor{black}{0.026}          & 0.0018                              & 0.074          \\
\rowcolor[HTML]{EFEFEF} 
\begin{tabular}[c]{@{}c@{}}Graph-RNN\\ (color)\end{tabular}                      & \textbf{\textcolor{black}{0.004}}                               & \textbf{\textcolor{black}{0.071} }           &    \textbf{\textcolor{black}{0.063}}                          &  \textbf{\textcolor{black}{0.219}}          & \textbf{\textcolor{black}{0.0003}}                         & \textbf{\textcolor{black}{0.025}}          &  \textbf{\textcolor{black}{0.0014}}                         & \textbf{\textcolor{black}{0.053}}           \\ \hline
\end{tabular}}
\caption{ Prediction error of PointRNN and Graph-RNN on the Synthetic Human Bodies and JPEG datasets }
    \label{table:results_bodies}
\end{table}
% colors in the Table just for me
\section{Conclusion}
\label{sec:conclusion}
This paper proposes end-to-end learning network to process dynamic PCs and make accurate predictions of future frames. We design a Graph-RNN cell that can leverage learned features, describing the local topology, to form spatio-temporal graphs, from where temporal correlations can be extracted. Experimental results demonstrate the network's ability to model  short and long-term motions while preserving the spatial structure.

\bibliographystyle{IEEE}
%\bibliography{strings,refs}
\bibliography{main}

%%%%%%%%%%%%%%
%SUPPLEMENTARY MATERIAL
\clearpage
\newpage

\include{supplement}

\end{document}

%% file: supplement.tex
\setcounter{table}{0}
\renewcommand{\thetable}{S\arabic{table}}%
\setcounter{figure}{0}
\renewcommand{\thefigure}{S\arabic{figure}}%
\setcounter{section}{0}

\begin{center}
\textbf{\large SPATIO-TEMPORAL GRAPH-RNN FOR POINT CLOUD PREDICTION} \\ SUPPLEMENTARY MATERIAL
\end{center}

\appendix
%\begin{flushleft}

This supplementary material provides additional details of the proposed framework.% as well additional visualizations of experimental results.

In Sec \ref{sec_A} we provide details on hierarchical structure. Sec \ref{sec_B} includes additional information of the datasets used in the experiments. Sec \ref{sec_C} provides implementation details of the architecture. Lastly Sec \ref{sec_D} provides visualization and analysis of additional experiments.

\section{Hierarchical structure details}
\label{sec_A}

In this paper, we proposed a hierarchical architecture, where before each Graph-RNN cell the point cloud and the associated components are down-sampled by a \emph{Sampling and Grouping} (SG) module. In a second phase, the point cloud is up-sampled to the original number of points \emph{State Propagation} (SP) module. 
The SG and SP modules were developed in the PointNET++~\cite{qi2017pointnet++} work. This section includes a description of the modules operations for the method proposed in this paper, for a more complete description we refer the reader to the original \cite{qi2017pointnet++} work.

\subsection{Sampling and Grouping}
The Sampling and Grouping  module takes a point cloud with $n$ points and uses the farthest point sampling (FPS) algorithm to sample $n'$ points. The sampled points are defined as centroids of local regions. Each region is composed of the $k$ closest neighborhood points to the centroid point. The features and states of the points in a region are max pooled into a single feature and state representation. This representation becomes the feature and the state of the centroid point. The SG module outputs the $n'$ sampled points and their updated feature and state.

\subsection{State Propagation}
In the SG modules, the original point set is down-sampled. However, in our prediction task, we want to obtain the point states for all the original points. The chosen solution is to propagate states from subsampled points $n' \times d_s $ to the original points $n \times d_s$.
To this end, for every down-sampling SG module, there is a corresponding up-sampling SP module, with a skip link connection between them as shown in Figure \ref{scheme}.
The SP module receives the target points we want to propagate the states into using skip connections, and interpolates the state's values $S$ of $n'$ points at coordinates of the $n$ points, using inverse distance weighted average based on k-nearest neighbors.
The interpolated states on \(n\) points are then concatenated with states from the SG module. The concatenation of both states is passed through an MLP to update every point state. The process is repeated until we have propagated states to the original set of points.\\

An additional advantage of the hierarchical architecture provided by the SG and SP modules is a reduction of computational power \cite{PointRNN}. This is a result of the reduced number of points processed in the layer after the down-sampling operations. 
Not only does the hierarchical architecture allow us to achieve better performance (more accurate predictions), informal evaluation during our experiments also confirmed a reduction of computation required.

\section{Dataset details}

This section provides details on point cloud datasets used in experiments.
\label{sec_B}
\subsection{ Moving MNIST Point Cloud}
The Moving MNIST Point Cloud dataset is a small, simple, easily trained dataset that can provide a basic understanding of the behavior of the network.

The dataset is created by converting the MNIST dataset of handwritten digits into moving point clouds. The sequences are generated using the process described in~\cite{PointRNN}. Each sequence consists of $20$ consecutive point clouds. Each point cloud contains one or two potentially overlapping handwritten digits moving inside a $64 \times 64$ area. Pixels whose brightness values (ranged from 0 to 255) are less than 16 are removed, and  128 points are randomly sampled for one digit and 256 points for two digits. Locations of the pixels are transformed to $(x,y)$ coordinates with the $z$-coordinate set to 0 for all points.

\subsection{Synthetic Human Bodies}
Open datasets for dynamic point clouds are limited, especially if interested in complex dynamic movements and not only body translation. Hence, we created synthetic data set of animated human bodies, similarly to \cite{Irene}.
We use the online service Mixamo \cite{Mixamo} to create multiple models of animated characters. Next, we used the 3D animation software Blender \cite{Blender} to render the animations and to extract one mesh per frame.
The mesh is converted to a high-resolution point cloud by randomly sampling $8000,0000$ points from the faces of the mesh. The point cloud is further downsampled to $4,000$ points using FPS to reduce memory and processing cost during experiments

The Human Bodies training dataset consists of $15$ character models each performing $20$ animations, for a total of $300$ sequences, we were careful to select a diverse group of activities. Each sequence contains $50$ frames, $12$  consecutive frames are randomly selected at each training step. The dataset is further augmented by using multiple sampling rates.

The test dataset consists of $5$ models denoted: Andromeda, James, Josh, Pete and Shae. All performing the same 9 activities: 'Big Jump', 'Climbing Up Wall', 'Entering Code','Jazz Dancing', 'Turn and Kick', 'Running', 'Stabbing', 'Walking Backwards', ' Walk with Rifle'. We again use different sampling rates to expand the dataset to a total of $152$ sequences.

\begin{figure}[h]
    \centering
    \subfigure[]{\includegraphics[width=0.44\columnwidth]{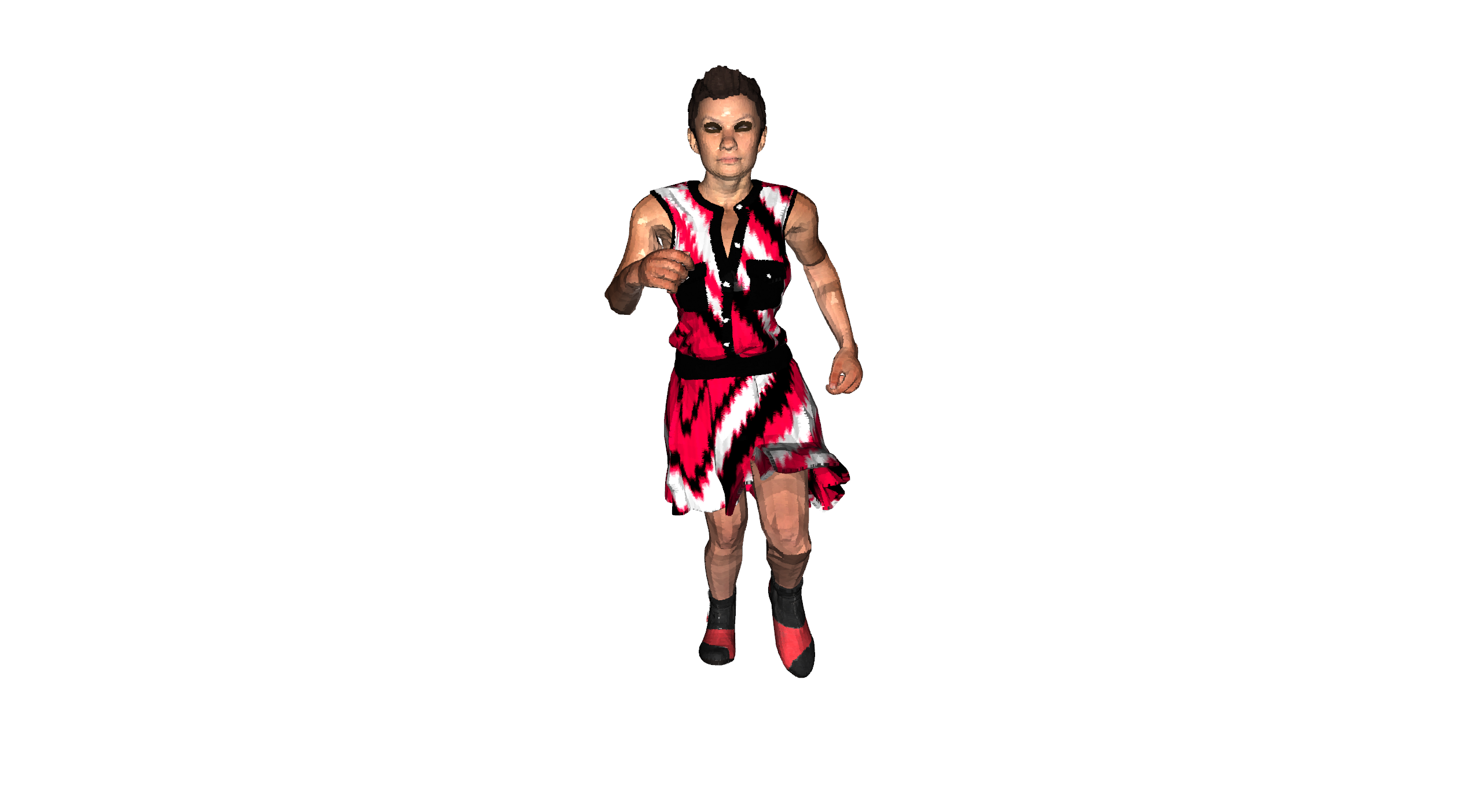}} 
    \subfigure[]{\includegraphics[width=0.44\columnwidth]{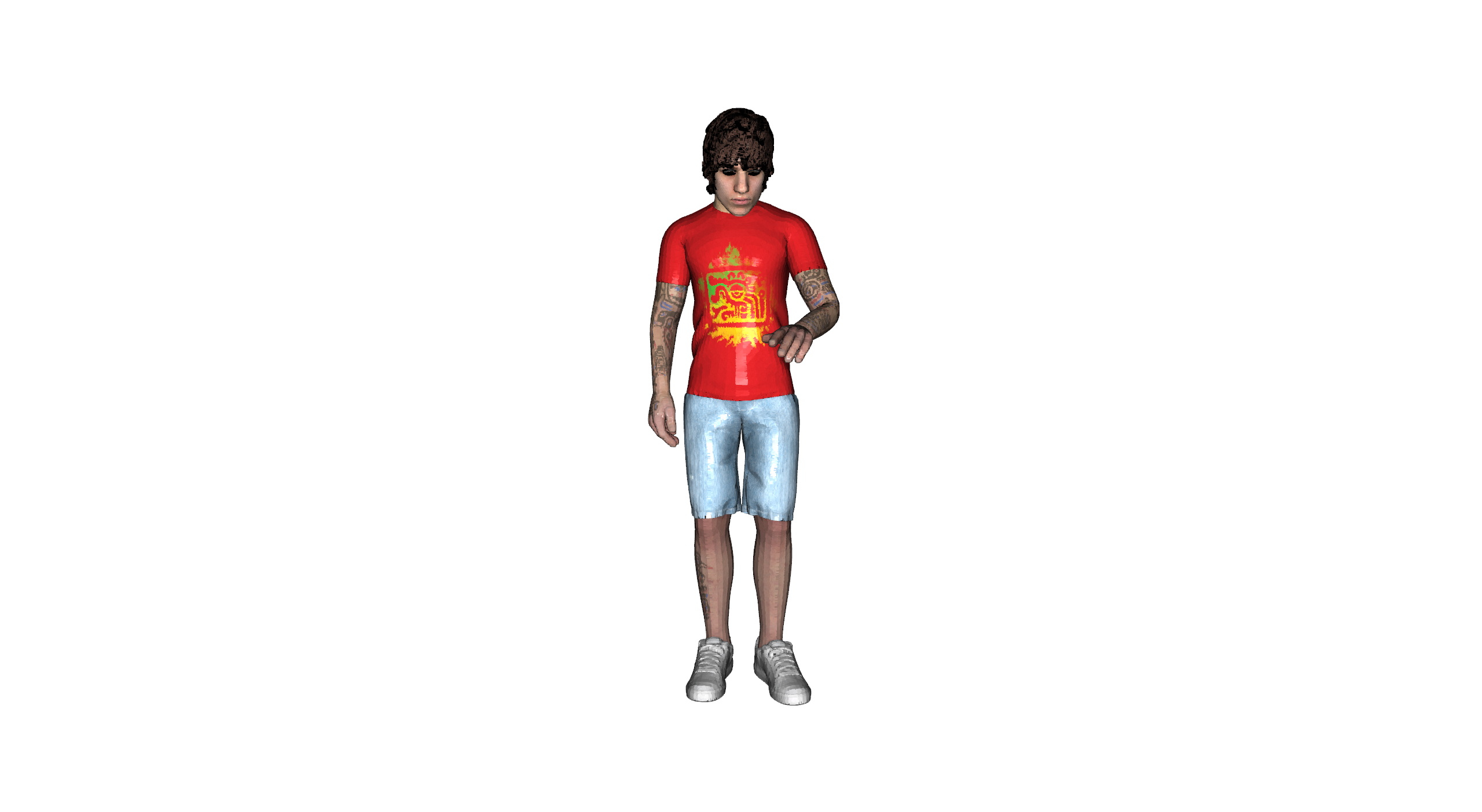}} 
    \subfigure[]{\includegraphics[width=0.44\columnwidth]{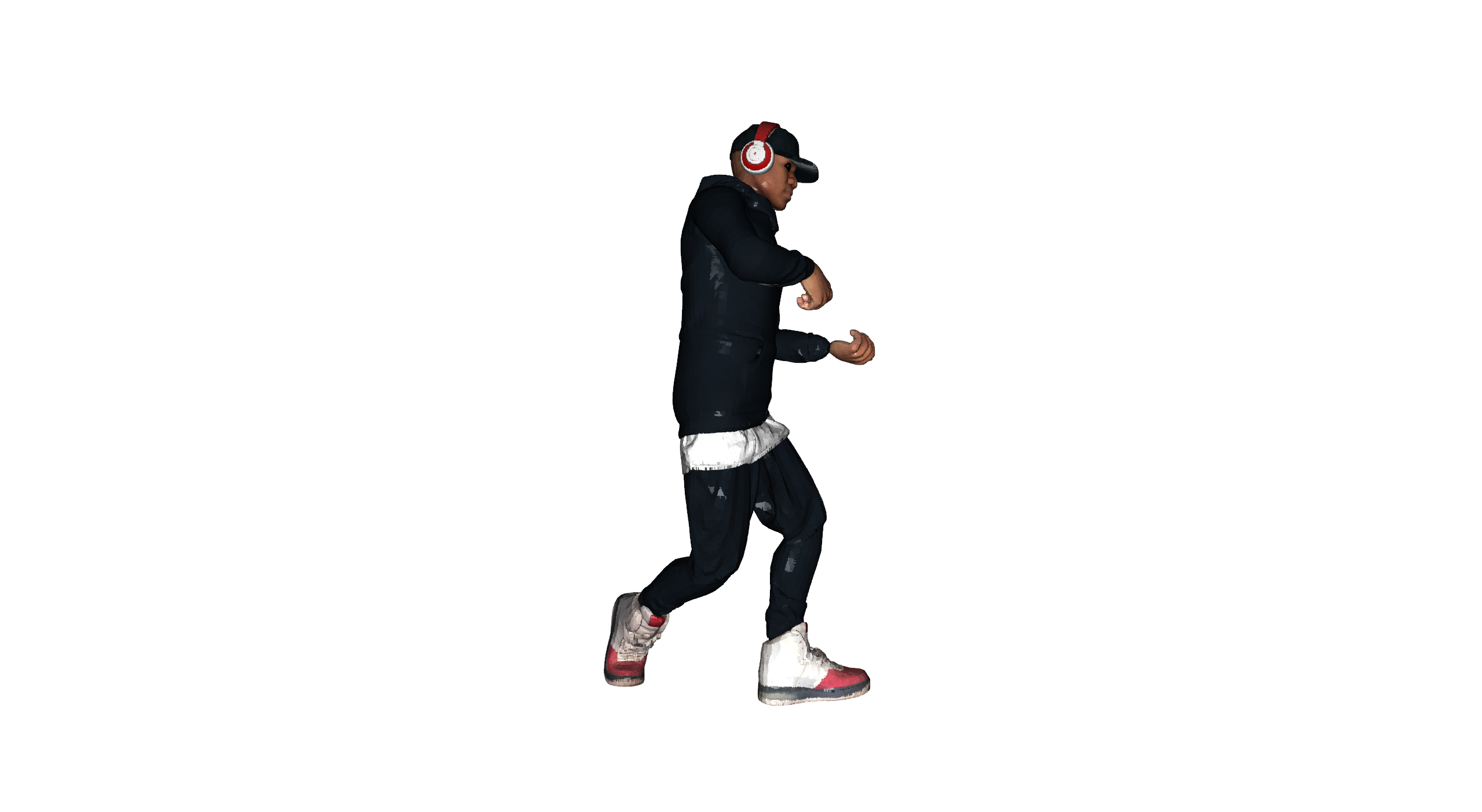}}
    \subfigure[]{\includegraphics[width=0.44\columnwidth]{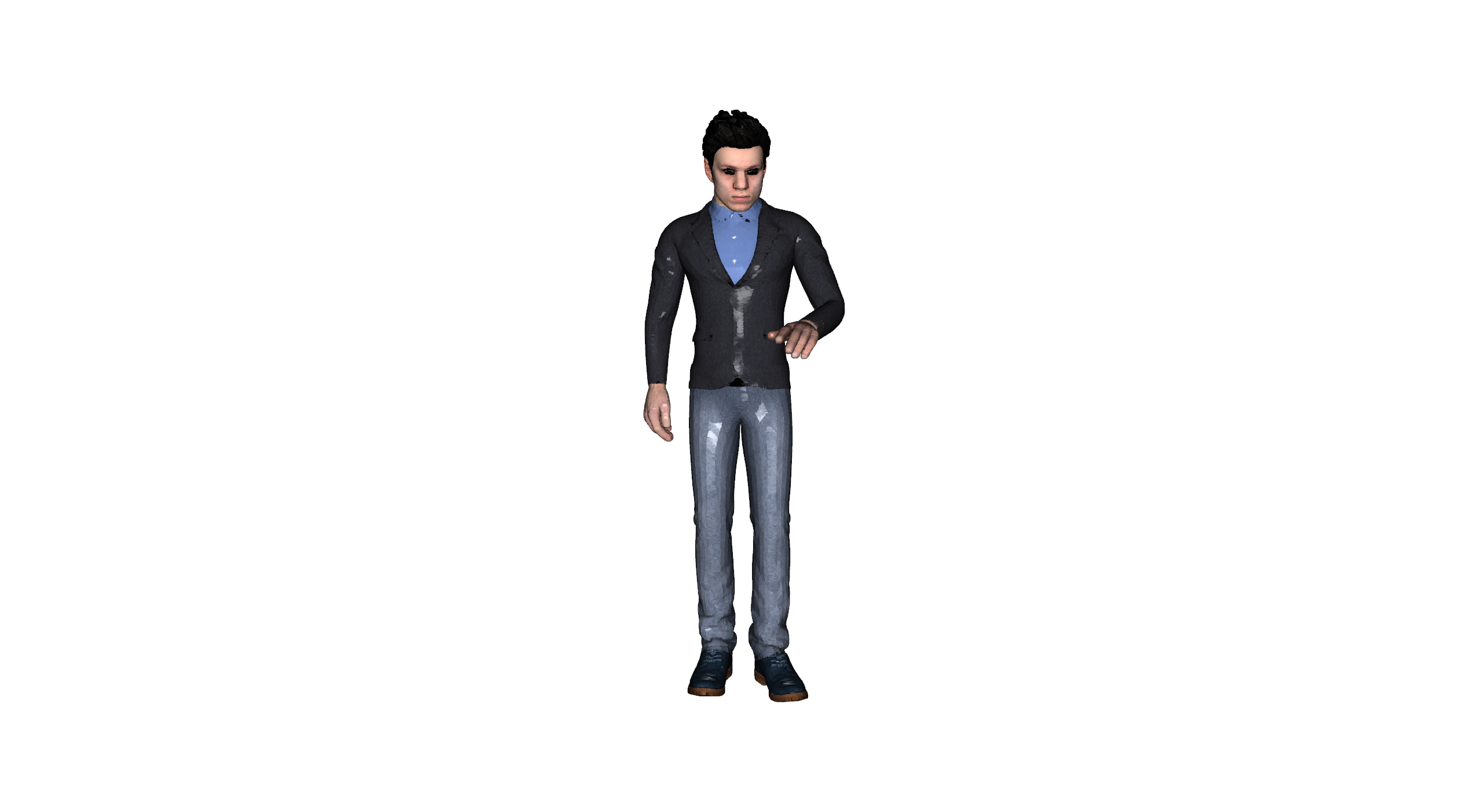}}
    \subfigure[]{\includegraphics[width=0.44\columnwidth]{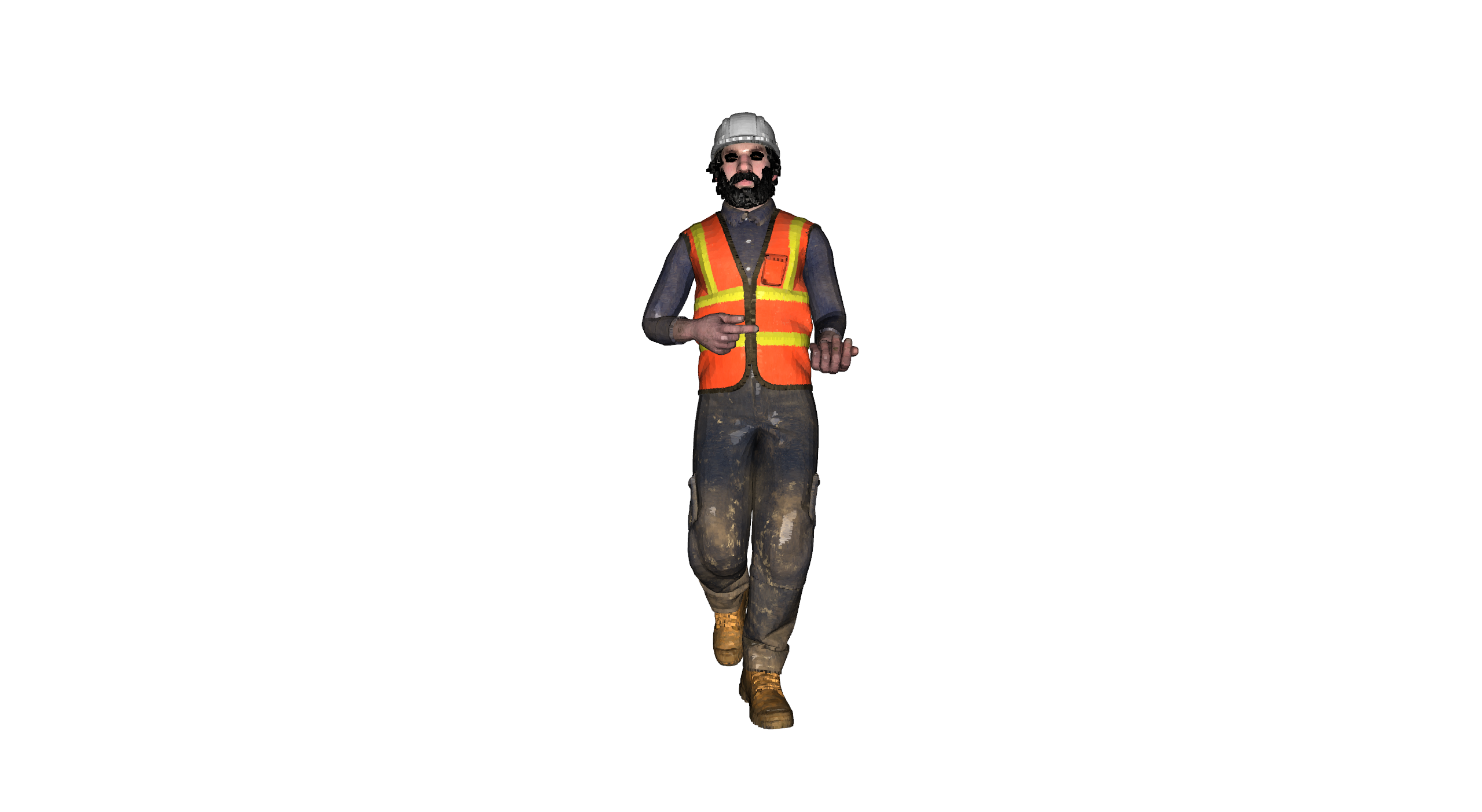}}
    \subfigure[]{\includegraphics[width=0.44\columnwidth]{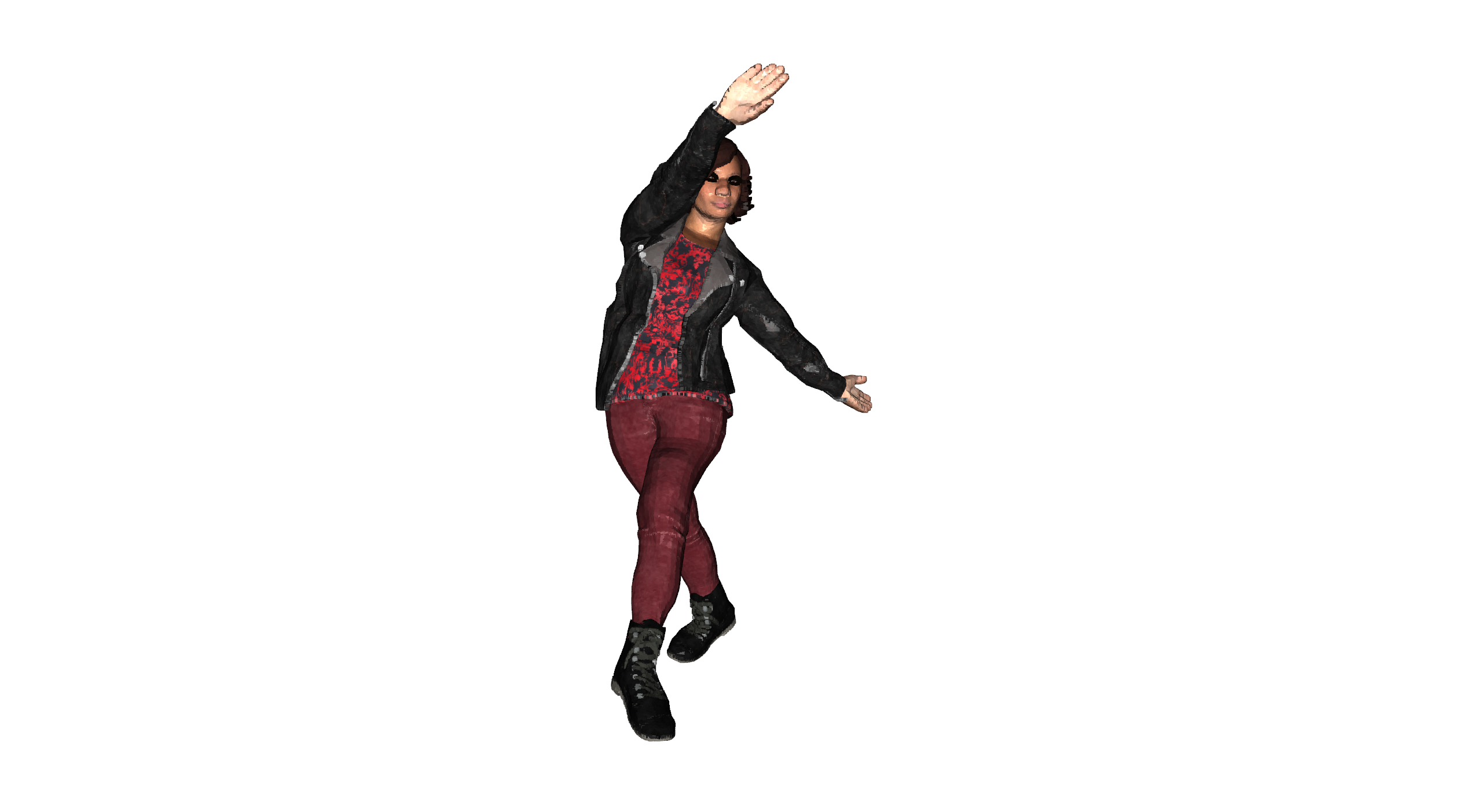}}
    \caption{Test characters: (a) Andromeda (b) Bryce (c) James  (d) Pete (f) Shae }
    \label{fig:foobar}
\end{figure}

\subsection{JPEG Pleno 8i Voxelized Full Bodies}
The dynamic voxelized point cloud sequences in this dataset are known as the 8i Voxelized Full Bodies (8iVFB). There are four sequences in the dataset, known as longdress, loot, redandblack, and soldier, pictured below. In each sequence, the full body of a human subject is captured by 42 RGB cameras configured in 14 clusters. The point clouds are originally high resolution with over $700,000$ points. The dataset is scaled by a factor of $0.0018$ and subsequently translated $(-0.37426165; -0.03379993; -0.29201281)$ to match the Human Bodies training data scale and general position. The data is then downsampled to 4,000 points using FPS.
\\
\begin{figure}[h]
    \centering
    \subfigure[]{\includegraphics[width=0.44\columnwidth]{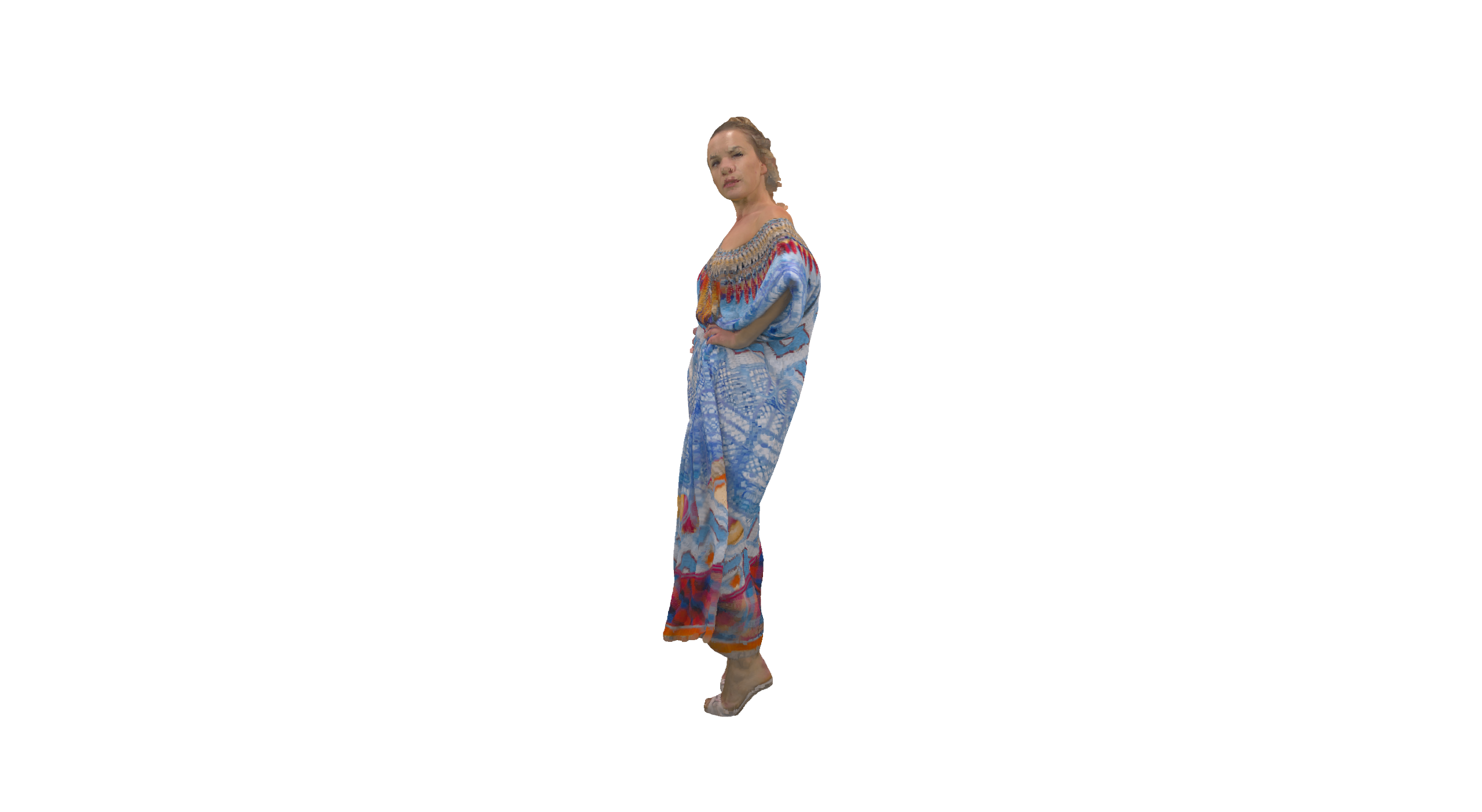}} 
    \subfigure[]{\includegraphics[width=0.44\columnwidth]{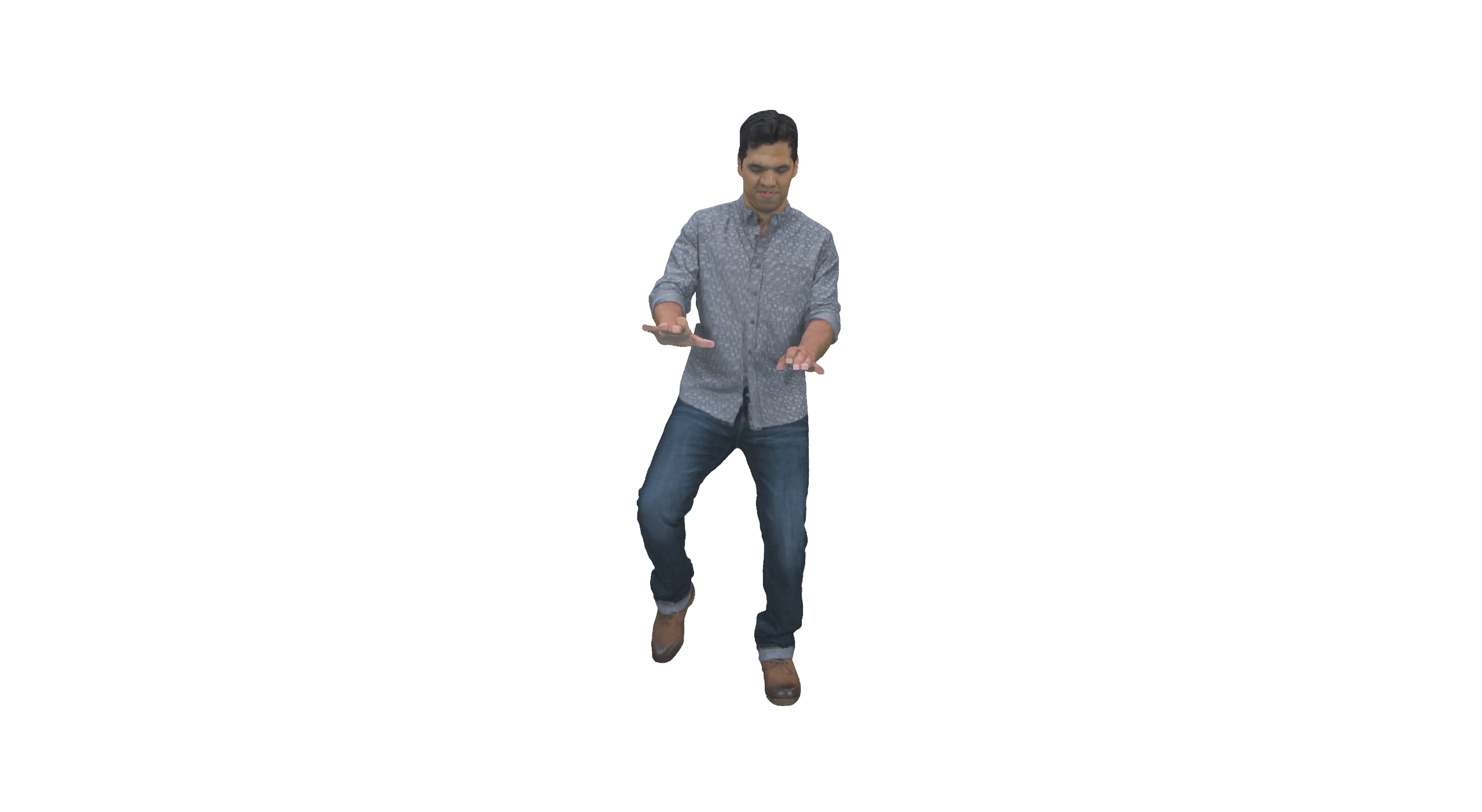}} 
    \subfigure[]{\includegraphics[width=0.44\columnwidth]{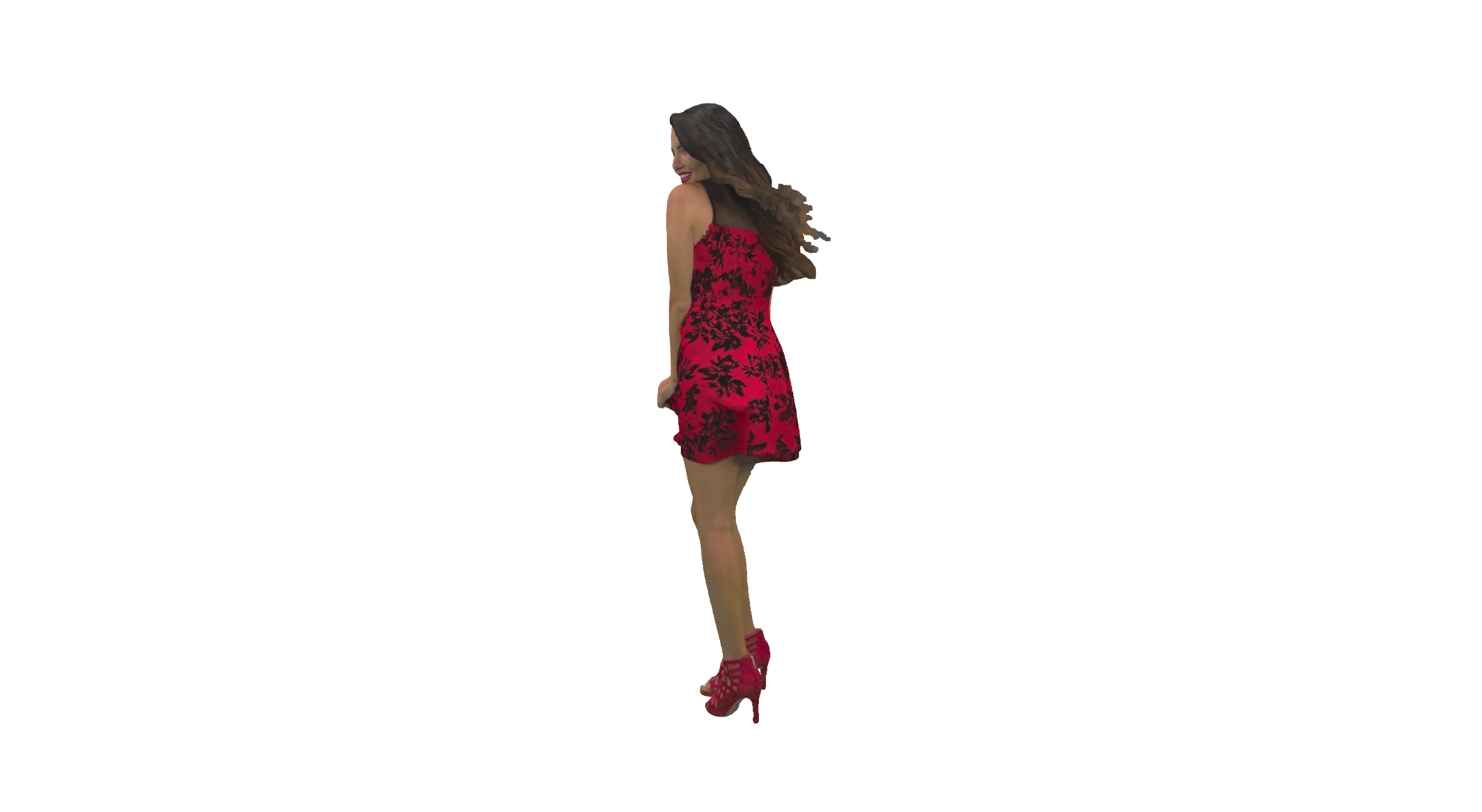}}
    \subfigure[]{\includegraphics[width=0.44\columnwidth]{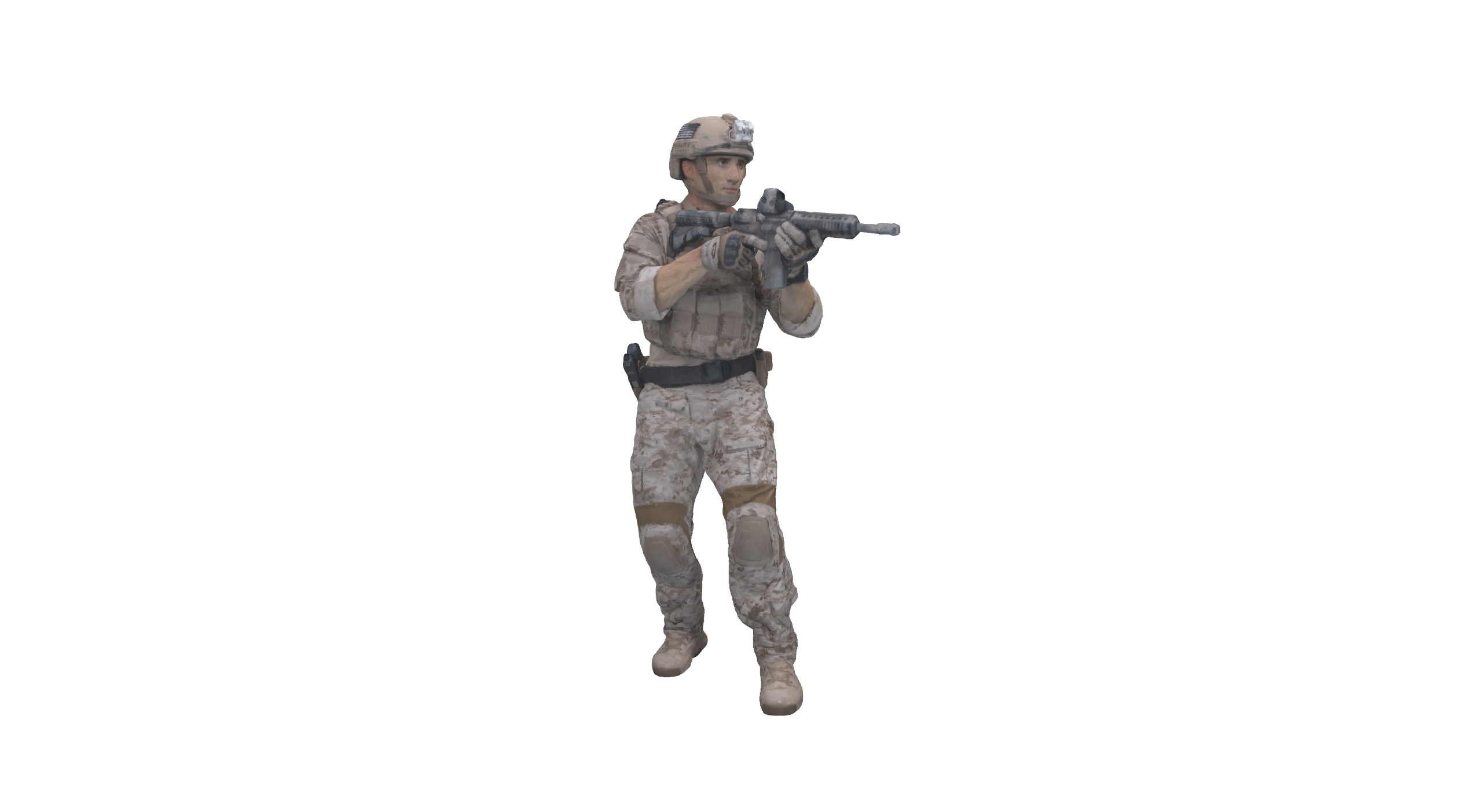}}

    \caption{JPEG Test characters: (a) longdress (b) loot (c) redandblack  (d) soldier }
    \label{fig:jpeg}
\end{figure}

FPS was chosen for the last downsample operations because it better coverage of the entire point cloud, compared with random sampling.
This dataset is only used for evaluation, of the models trained with the Synthetic Human Bodies dataset.

\section{Implementation details}
\label{sec_C}
This section describes the parameters and specifications  of the proposed framework.
\subsection{Training details}
The models are trained using the Adam optimizer,  with a learning rate of $10^{-5}$ for all datasets. The models trained with the MNIST dataset are trained for $200,000$ interactions with a batch size of 32. For the Synthetic Human Bodies dataset, the PointRNN and Graph-RNN models a batch size of 4 is set and trained for $200,000$ interaction in long-term prediction task and for  $150,000$ interaction in short-term prediction task. The Graph-RNN (color) model that considers point clouds with color is trained for $200,000$ interaction for both tasks with a batch size of 2. For all models, the gradients are clipped in range~$[5,5]$.

%Due the huge size of the 'Synthetic Human Bodies' Dataset, the training data can not be all loaded to memory during training. We design a system that load at random between 50 to 100 sequences to the program at the time. After 50 training interactions (Epochs) the the data is reloaded.

%%%%%%%%%%%%%%%%%%%%%
\subsection{Architecture Specifications}
This section provides the specification for each of the main components:
Sampling and Grouping (SG); The  Graph neural network (GNN) for features learning; Graph-RNN cells; States propagation (SP); Final Fully connected layer (FC);

\hfill

The Graph-RNN model is implemented the same way for the MNIST dataset and the Synthetic Human Bodies. For the MNIST dataset, we compare Graph-RNN results with the original PointRNN results ($k$-nn Model). 
However, since the original PointRNN paper \cite{PointRNN} did not perform experiments on the Synthetic Human Bodies dataset, we choose the $k$-values and dimensions to adapt the PointRNN framework to the dataset. To have a fair comparison between our proposed Graph-RNN and Point-RNN, we tried to keep the frameworks as similar as possible while preserving the design choices of each one. 

The architecture specifications of both Graph-RNN and PointRNN are displayed in Table S1.

\begin{table}[h]
\centering
\resizebox{1\columnwidth}{!}{%
\begin{tabular}{|cc|ccc|ccc|}
\hline
\multicolumn{8}{|c|}{Specifications}                                                                                                                                                                         \\ \hline
\multicolumn{2}{|c|}{\begin{tabular}[c]{@{}c@{}}number of\\ output points\end{tabular}} & \multicolumn{3}{c|}{Graph-RNN model}                     & \multicolumn{3}{c|}{Point-RNN model}                    \\ \hline
\textit{hierarchical}                          & \textit{basic}                         & Components        & \textit{k} & \textit{ouput channels} & Components       & \textit{k} & \textit{ouput channels} \\ \hline
\rowcolor[HTML]{EFEFEF} 
\textit{n/2}                                     & \textit{-}                             & SG                & 4          & -                       & SG               & 4          & -                       \\
\textit{n/2}                                   & \textit{n}                             & GNN layer 1       & 16         & 64                      & -                & -          & -                       \\
\textit{n/2}                                   & \textit{n}                             & GNN layer 2       & 16         & 128                     & -                & -          & -                       \\
\textit{n/2}                                   & \textit{n}                             & GNN layer 3       & 8          & 128                     & -                & -          & -                       \\
\rowcolor[HTML]{EFEFEF} 
\textit{n/2}                                   & \textit{n}                             & Graph-RNN  cell 1 & 8          & 256                     & PointRNN  cell 1 & 24         & 256                     \\
\textit{n/4}                                   & \textit{-}                             & SG                & 4          & -                       & SG               & 4          & -                       \\
\rowcolor[HTML]{EFEFEF} 
\textit{n/4}                                   & \textit{n}                             & Graph-RNN  cell 2 & 8          & 256                     & PointRNN cell 2  & 16         & 256                     \\
\textit{n/8}                                   & \textit{-}                             & SG                & 4          & -                       & SG               & 4          & -                       \\
\rowcolor[HTML]{EFEFEF} 
\textit{n/8}                                   & \textit{n}                             & Graph-RNN  cell 3 & 8          & 256                     & PointRNN cell 3  & 8          & 256                     \\
\textit{n/4}                                   & \textit{-}                             & SP 1              & -          & 256                     & SP 1             & -          & 256                     \\
\textit{n/2}                                   & \textit{-}                             & SP 2              & -          & 256                     & SP 2             & -          & 256                     \\
\textit{n}                                   & \textit{-}                             & SP 3              & -          & 256                     & SP 3             & -          & 256                     \\
\rowcolor[HTML]{EFEFEF} 
\textit{n}                                     & \textit{n}                             & FC 1              & -          & 126                     & FC1              & -          & 128                     \\
\rowcolor[HTML]{EFEFEF} 
\textit{n}                                     & \textit{n}                             & FC 2              & -          & 3                       & FC 2             & -          & 3                       \\ \hline
\textit{-}                                     & \textit{-}                             & \multicolumn{3}{c|}{Graph-RNN (color) model}             & -                & -          & -                       \\
\textit{n}                                     & \textit{n}                             & FC color 1        & -          & 126                     & -                & -          & -                       \\
\textit{n}                                     & \textit{n}                             & FC color 2        & -          & 3                       & -                & -          & -                       \\ \hline
\end{tabular}
}
\caption{Architecture specifications. Each component is described by tree attributes, i.e number of output points, number of neighborhoods ($k$) and number of output channel.}
\end{table}

For all the models, the final fully connected (FC) layer is implemented in fact by two fully connected layers FC1 and FC2. The FC1  and FC2 layers process the states to predict the geometry displacement.

The Graph-RNN (color) model that takes color as input has two additional fully connected layers (FC1 color and FC2 color). Similar to the FC for points, the FC (color) will take the states as input and predict the color displacement.
The assumption that the color of the points does not change during the movement, while mostly correct in the case of synthetically generated data, can not be applied in real-world data. The point's color can change due to lighting conditions, or in extremes cases, scene objects can transform, or be replaced by new ones. The color prediction was not a priority in this work. The FC (color) does not affect the loss function and has no impact on the overall method. Nevertheless, in long-term prediction evaluation, we disregard the prediction of color displacement and consider instead a null color displacement, meaning all the points keep the same color from frame to frame.
In the future, we intend to explore color prediction, by including a color evaluation metric in the loss function.

\section{Extra Results visualization}
\label{sec_D}
This section presents visualization examples of prediction on Synthetic Human Bodies and JPEG datasets. Long-term prediction examples are depicted on the right side of the next page and short-term prediction examples on left.

\hfill
%%%% long-term description

Long-term prediction is a very challenging task. In the MNIST dataset, the moving digits perform simple translation movements. The Graph-RNN can effectively model these simple motions over a long period, and make a long-term prediction (10 frames). On the other hand, Synthetic Human Bodies perform more complex activities (e.g dancing, climbing, running). Since these activities are composed of irregular movements, with sudden motion changes, they are incredibly difficult to predict in the long-term. 

Figure S3 and Figure S4, show the difficulty in long-term prediction in human activities. Both the PointRNN and Graph-RNN have trouble at preserving the spatial structure over time. This was expected since in long-term prediction the error in each predicted frame is propagated and amplified for each subsequent prediction.
While from Figures S3 and S4, it would appear the PointRNN is better at preserving the spatial structure, this is because the PointRNN failed to capture the correct motion. The Graph-RNN, on the other hand, correctly captured the general motion, losing however some of the point cloud shape as a result. We can conclude that a better prediction comes with the risk of higher deformation. The Point-RNN achieved a smaller prediction error in the long-term prediction of the Human Bodies dataset, by making more conservative predictions (low movement prediction). In contrast, the Graph-RNN makes more accurate predictions, however in the cases the predicted motion is wrong, the error is amplified, resulting in a worse average performance.

Figure S6 and S7 display short-term prediction examples. All the models show very similar visual results and can make accurate predictions. The Graph-RNN superior performance can be observed in the small details like in the knee and foot of the model.

Figures S5 and S8 show examples on the JPEG Human Bodies dataset. The point clouds sequence from the JPEG dataset has very little movement, confirmed by the good performance of the Copy Last Input model in both short and long-term prediction. While the Graph-RNN achieved a smaller prediction error in the chosen evaluation metrics, it is difficult to see the performance gains by looking at the prediction visualizations.

\clearpage
\newpage
\begin{figure}[h]
    \centering
    \includegraphics[height=0.24\textwidth]{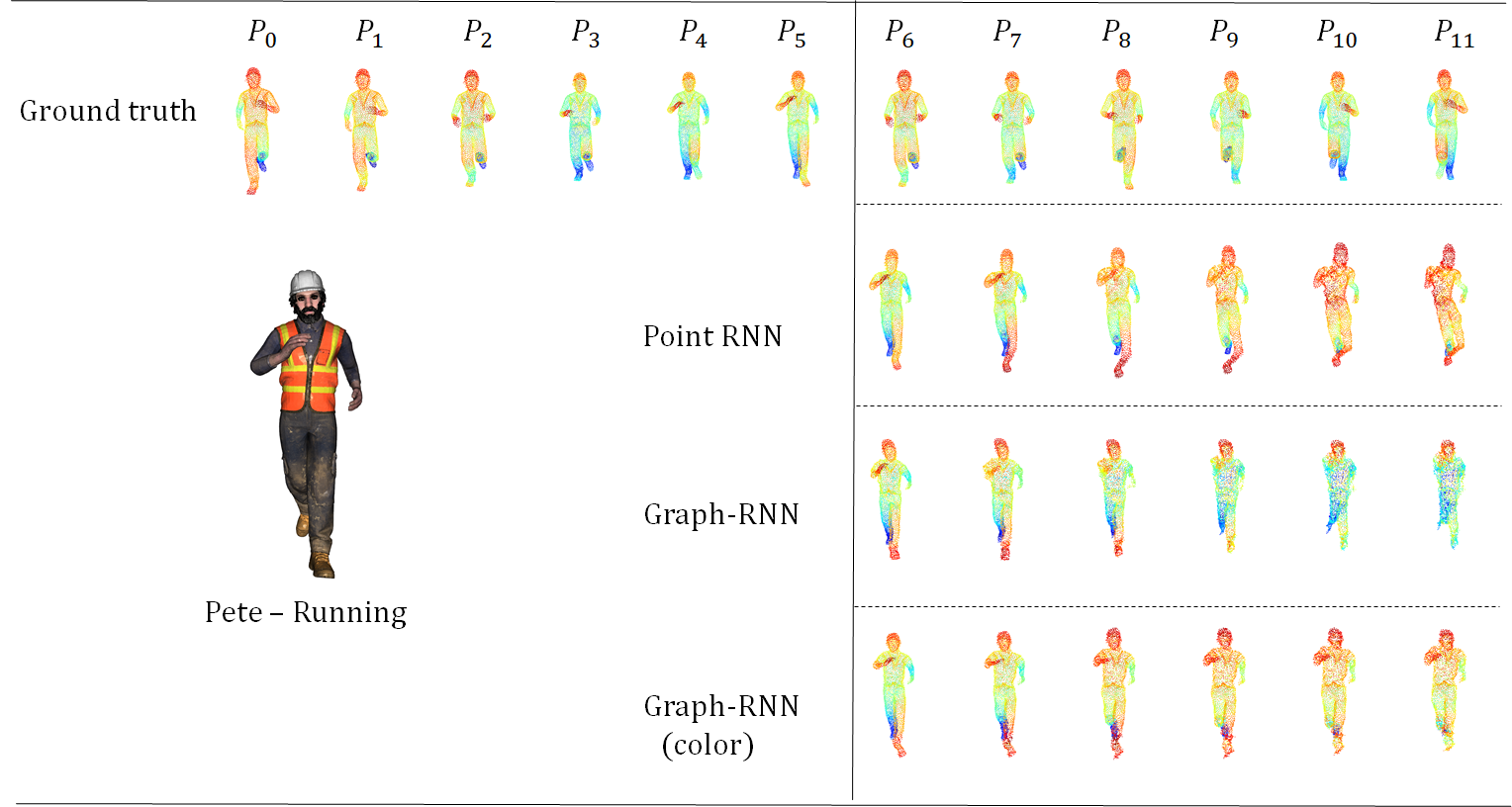}
    \caption{Short-term prediction on 'Pete Running' sequence}
\end{figure}

\begin{figure}[h]
    \centering
    \includegraphics[height=0.24\textwidth]{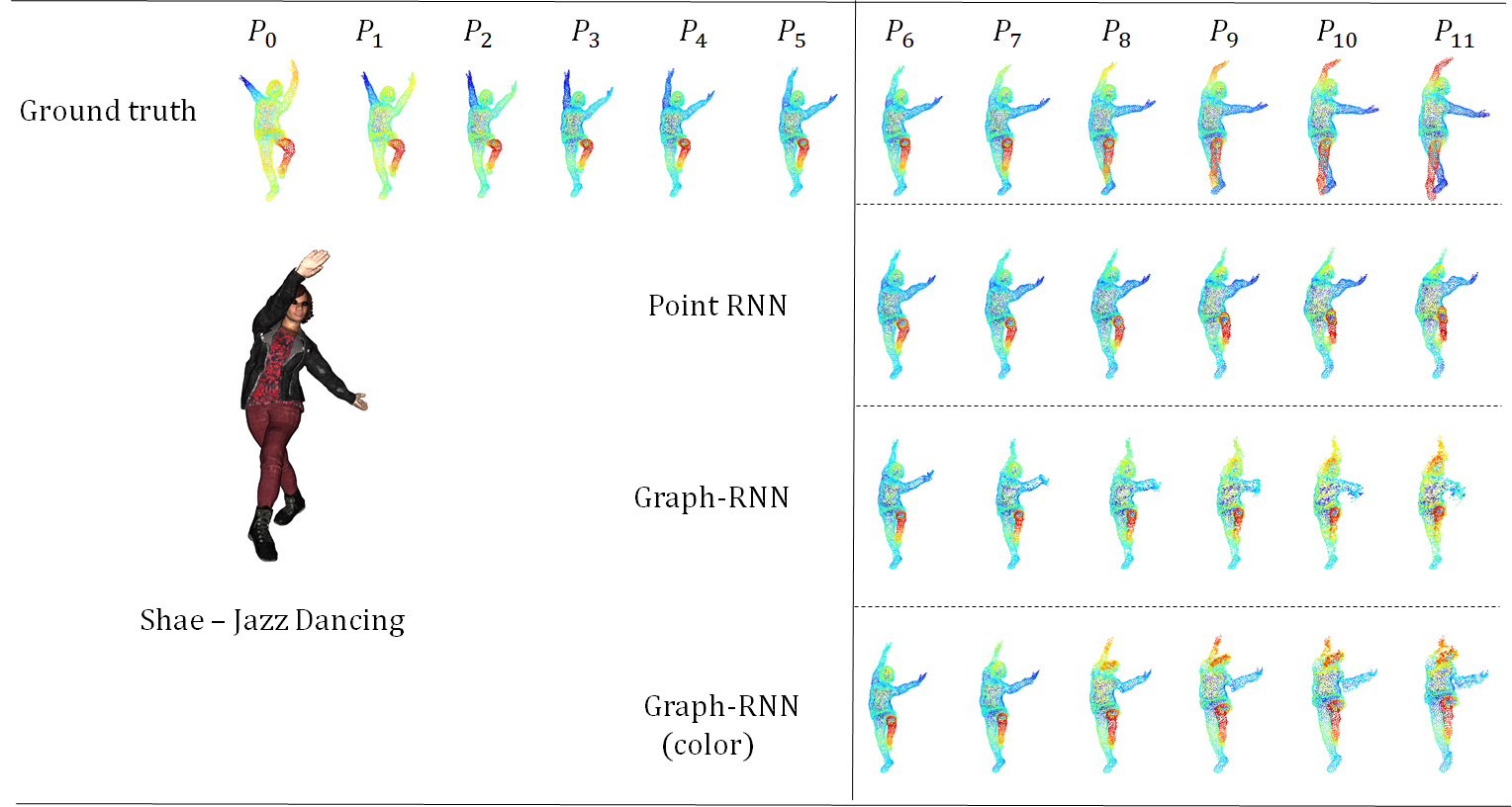}
    \caption{Short-term prediction on 'Shae Jazz Dancing' sequence}
\end{figure}

\begin{figure}[h]
    \centering
    \includegraphics[height=0.24\textwidth]{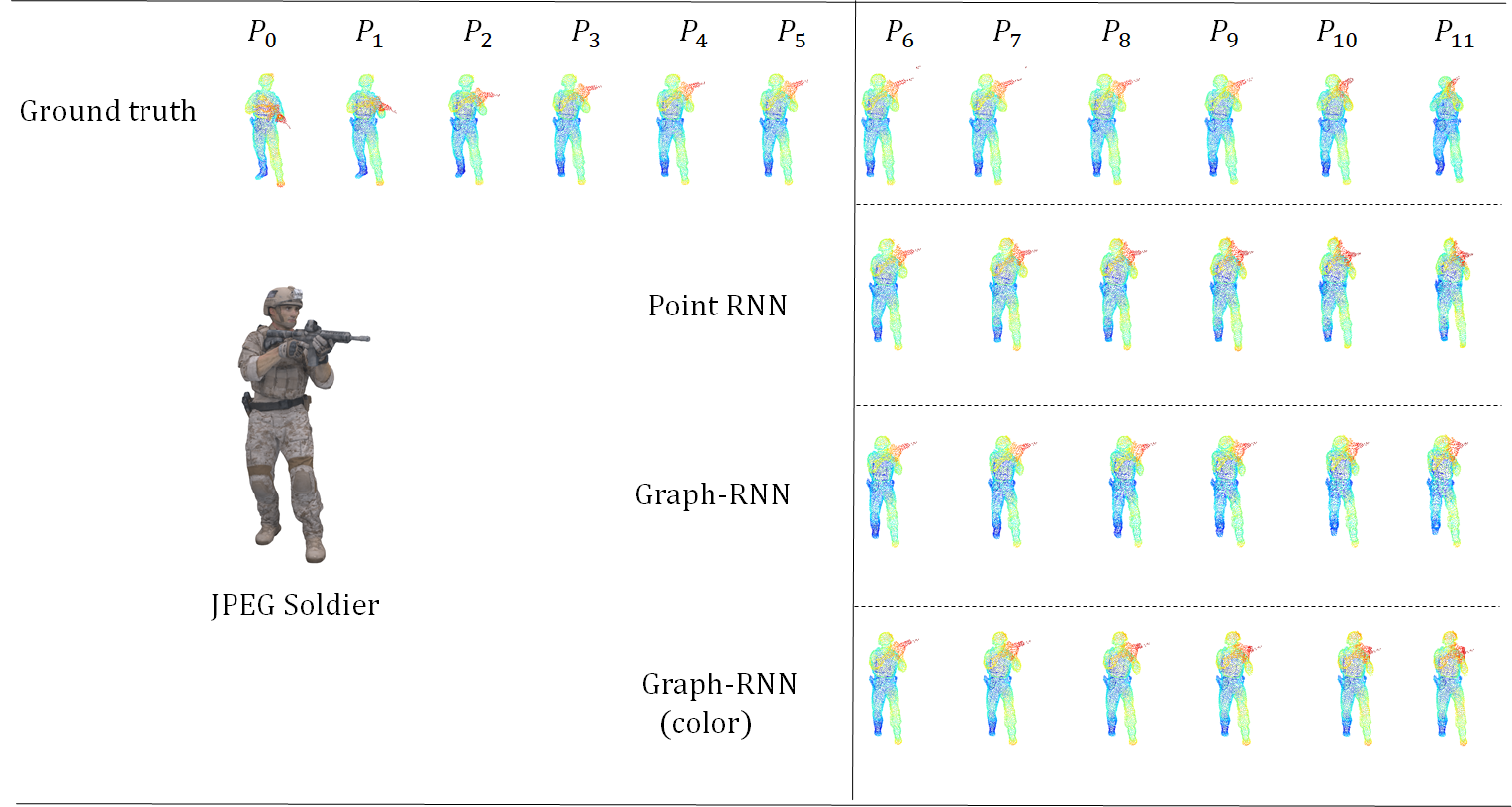}
    \caption{Short-term prediction on 'JPEG Soldier' sequence}
\end{figure}

\newpage

\begin{figure}[h]
    \centering
    \includegraphics[height= 0.24\textwidth]{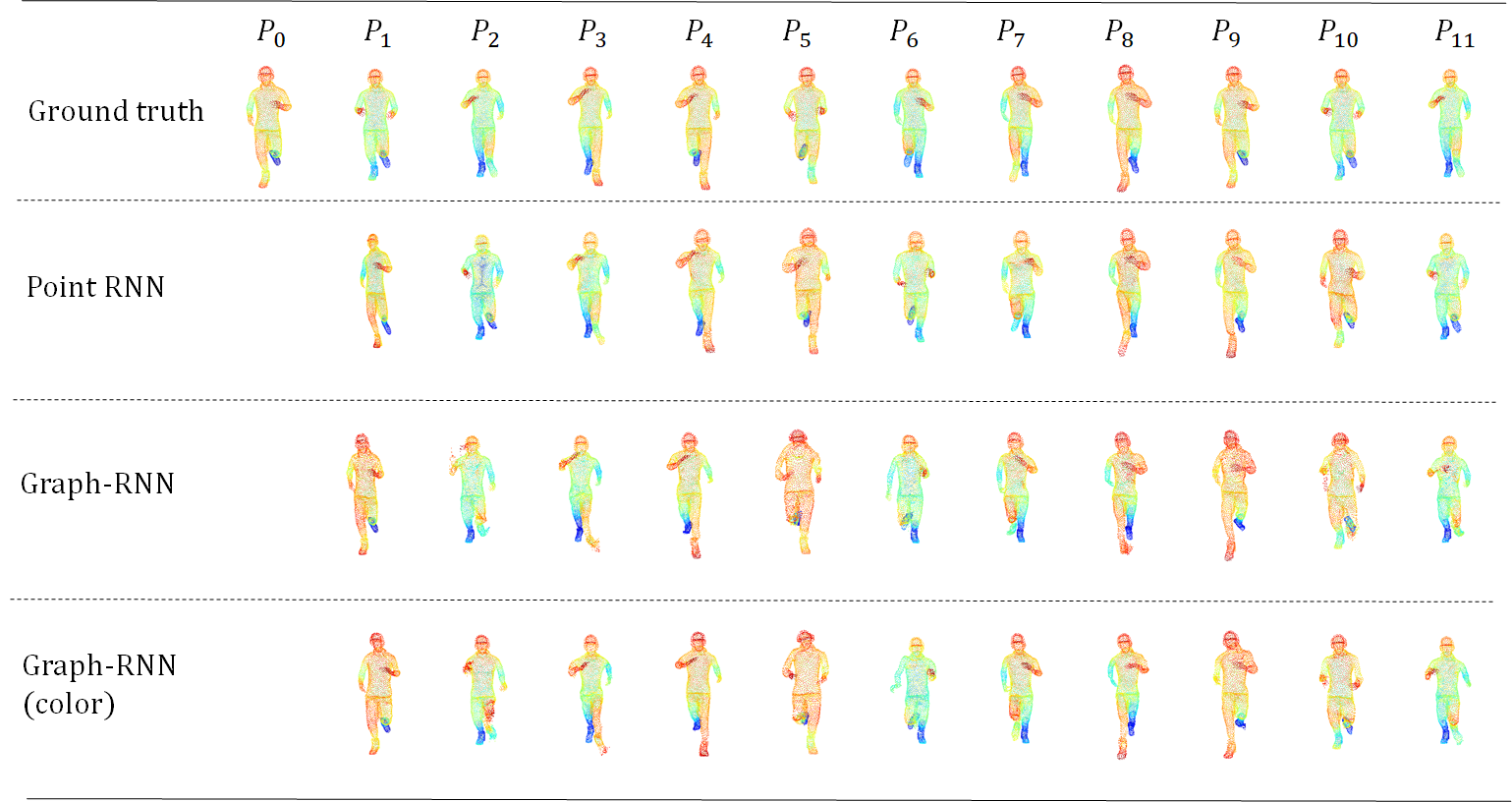}
    \caption{Short-term prediction on 'James Running' sequence}
\end{figure}

\begin{figure}[h]
    \centering
    \includegraphics[height=0.24\textwidth]{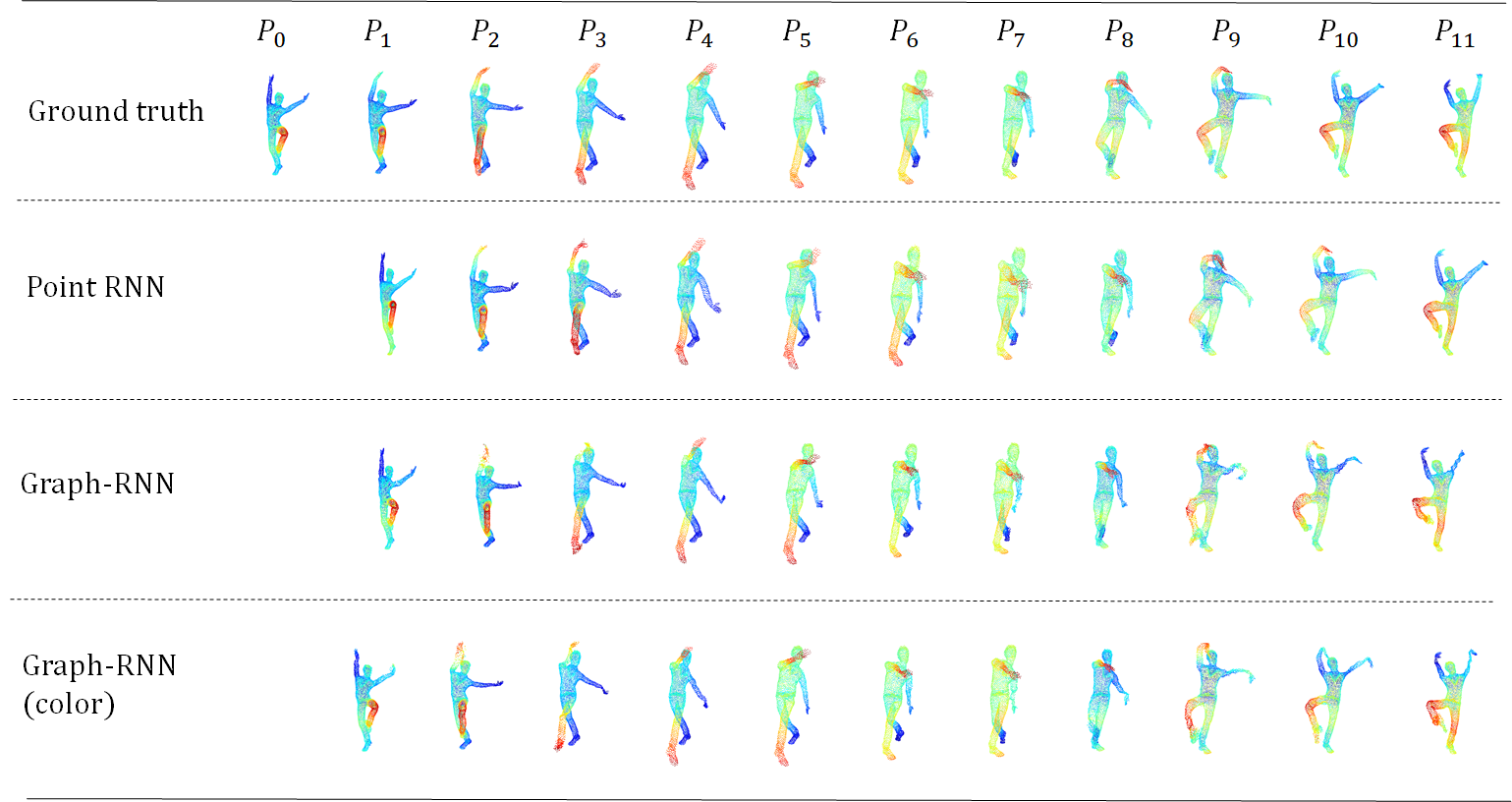}
    \caption{Short-term prediction on 'Josh Jazz Dancing' sequence}
\end{figure}

\begin{figure}[h]
    \centering
    \includegraphics[height=0.24\textwidth]{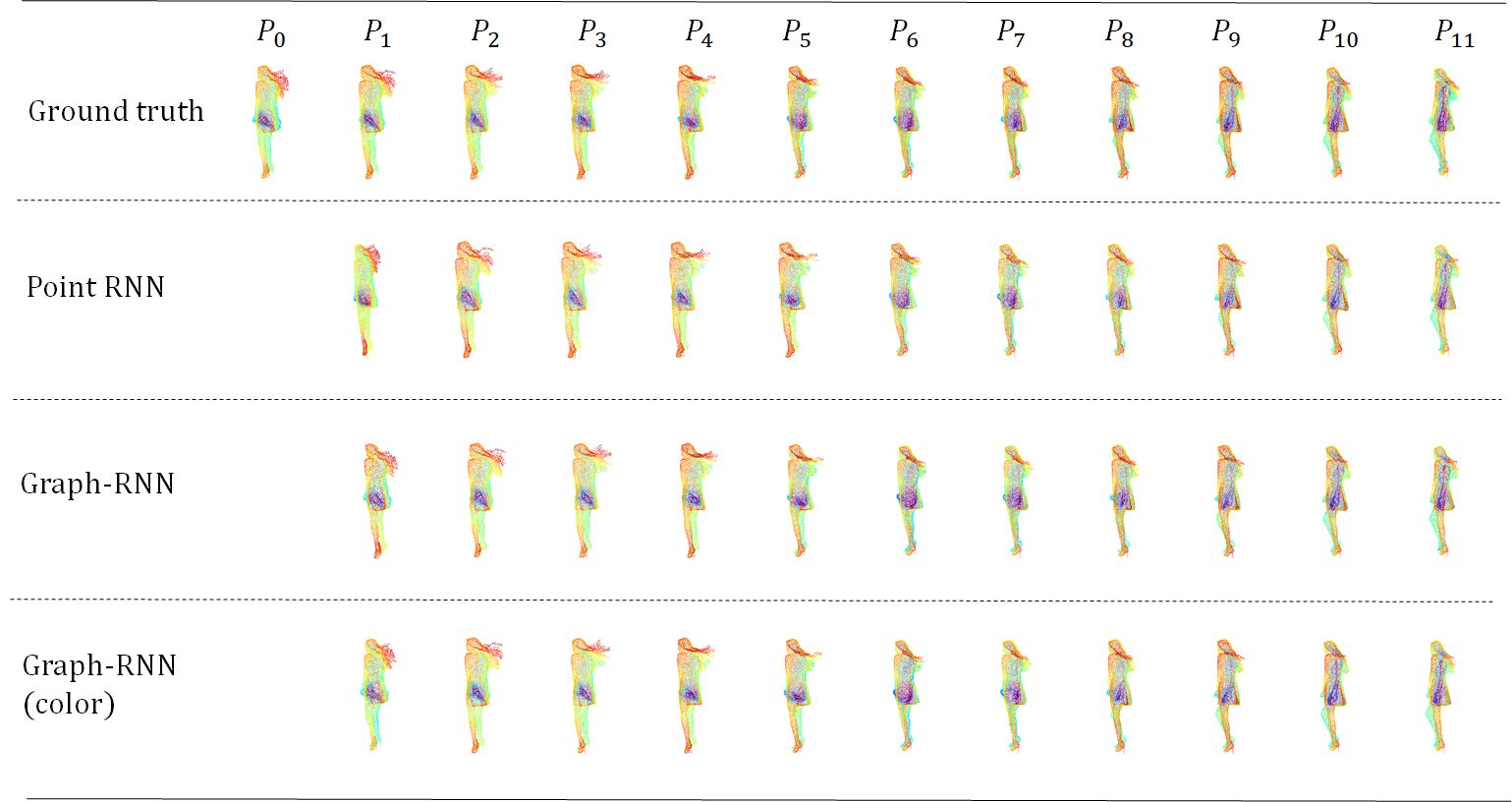}
    \caption{Short-term prediction on 'JPEG Redandblackdress" sequence}
\end{figure}

%\end{flushleft}